\documentclass[journal ]{new-aiaa}
\usepackage[utf8]{inputenc}
\usepackage{textcomp}
\usepackage{multirow}
\usepackage{graphicx}
\usepackage{amsmath}
\usepackage[version=4]{mhchem}
\usepackage{siunitx}
\usepackage{longtable,tabularx}
\title{Conflict Resolution under Degraded Surveillance in Air Corridors Using Multi-Agent Reinforcement Learning}

\author{
Esrat Farhana Dulia\textsuperscript{1}\footnote{Graduate Research Assistant, College of Aeronautics and Engineering, edulia@kent.edu.},  
Syed Arbab Mohd Shihab\textsuperscript{1}\footnote{Assistant Professor, College of Aeronautics and Engineering, sshihab@kent.edu.}, 
Caleb Adams\textsuperscript{1}\footnote{Graduate Research Assistant, College of Aeronautics and Engineering, cadams68@kent.edu.}, and
Ruben Del Rosario\textsuperscript{1}\footnote{Director, Center for Advanced Air Mobility and Professor, College of Aeronautics and Engineering, rdelros1@kent.edu.}
}

\affil{
\textsuperscript{1}Kent State University, Kent, OH, USA 44242 \\
}

\begin{document}

\maketitle

\begin{abstract}
Safe Advanced Air Mobility operations require aircraft to maintain separation when surveillance information is noisy, delayed, incomplete, or temporarily unavailable. This study develops a Deep Q-Network-based Multi-Agent Reinforcement Learning framework for decentralized conflict resolution among heterogeneous small unmanned aerial vehicles and electric vertical takeoff and landing aircraft operating within a structured three-dimensional corridor. Separate policies are trained for the two aircraft categories using local observations and a 14-action space that includes maintaining course, turning, vertical maneuvering, landing, and speed control. The simulation incorporates aircraft-specific dynamics, energy use, corridor constraints, observation noise, communication delay, information dropout, wind disturbance, actuator uncertainty, and model uncertainty. The trained policies are evaluated across 90 combinations of traffic density and minimum separation thresholds. Loss-of-separation frequency and duration generally increase with traffic density and separation requirements, although most events are resolved within 1s. Under safe conditions, agents maintain their motion approximately 79\% of the time. During conflicts, turning accounts for 33\% of actions, followed by maintaining motion at 29\%, speed control at 25\%, and vertical maneuvers at 13\%. Six Pareto-optimal configurations reveal trade-offs between safety and corridor capacity. The framework supports the simulation-based evaluation of safer AAM conflict-resolution strategies under degraded surveillance conditions.

\end{abstract}

\section{Introduction}

\subsection{AAM Operations and Degraded Surveillance}


AAM is expected to provide substantial economic, social, and environmental benefits by improving transportation accessibility, reducing travel time, supporting emergency response, and enabling more sustainable mobility services \cite{Dulia2021AAM,DelRosario2021Infrastructure,open_framework_standards,Calhoun2023OpenFramework}. To enable these operations, considerable research has been conducted on key aspects of the AAM ecosystem, including air traffic management \cite{FAA2020}, concepts of operation \cite{thipphavong2018urban}, market analysis \cite{hasan2018urban}, infrastructure investment planning \cite{dulia2024building,dulia2024negotiate}, operational planning \cite{Dulia2025LowNoiseUAV,dulia2025dynamic}, and supply chain planning \cite{dulia2024integrated}. Despite these advances, ensuring safe aircraft separation under degraded surveillance conditions remains an important challenge for large-scale AAM deployment. Integrating these aircraft safely into a shared airspace system will require communication, navigation, surveillance, automation, information exchange, and traffic-management services \cite{faa2023uamconops,stouffer2020cns}. Among these services, surveillance will be essential because aircraft will require accurate and timely information about nearby traffic to estimate relative motion, detect potential conflicts, and determine whether a maneuver is necessary \cite{kochenderfer2012acasx}. Under normal operating conditions, this information is expected to be obtained from a combination of ground-based and onboard surveillance sources. An information clearinghouse is expected to integrate, process, and distribute the collected data to aircraft and traffic-management services \cite{dulia2024surveillance}. By combining information from multiple sources, the surveillance system is expected to provide broader coverage and more complete estimates of aircraft position, speed, heading, and trajectory.

This level of surveillance support will depend on the continued operation of ground-based sensors, communication links, the information clearinghouse, power systems, and other supporting infrastructure. Failures or disruptions in these components can reduce the availability and reliability of ground-based traffic information. As ground-based information becomes limited or unavailable, aircraft will need to rely more heavily on onboard surveillance to observe nearby traffic. Onboard sensors, however, provide a more restricted local view and are affected by measurement errors, limited sensing range, line-of-sight obstruction, environmental conditions, signal interference, communication delay, information dropout, missed detections, false alarms, and sensor-specific accuracy limitations. Consequently, the combination of ground-system disruption and onboard-sensor uncertainty will reduce the accuracy, timeliness, completeness, and availability of the surveillance information accessible to an aircraft \cite{dulia20263r,dulia2024surveillance,hespanha2007networked,sinopoli2004kalman}. In this degraded-surveillance environment, an aircraft can receive inaccurate position estimates, outdated movement information, incomplete observations of surrounding traffic, or no observation during part of an encounter. These limitations can lead to incorrect separation estimates, reduce the time available for a response, and prevent timely conflict detection \cite{sinopoli2004kalman,zhang2020robustdrl,agarwal2021delayedobservations}. An onboard conflict-resolution capability is therefore needed to support safe maneuvering when surveillance information is uncertain, delayed, incomplete, or temporarily unavailable.

\subsection{Multi-Aircraft Conflict Resolution}

In addition to surveillance limitations, conflict resolution in AAM is complicated by the presence of heterogeneous aircraft. Small UAVs and eVTOL aircraft operate with different speeds, altitude ranges, maneuvering capabilities, mission requirements, and energy limitations \cite{thipphavong2018uam,bauranov2021airspace}. When these aircraft operate within the same transportation network, their operational differences influence how quickly conflicts develop, how much response time is available, and which maneuvers can be performed safely. Maintaining safe operations therefore requires adequate separation, defined operating areas, and timely responses to surrounding traffic \cite{kuchar2000review}.

The problem becomes more complex when several aircraft interact simultaneously because conflict resolution is not an independent decision for each aircraft. A maneuver selected by one aircraft changes its relative position and motion with respect to nearby traffic and can reduce the maneuvering options available to other aircraft \cite{lai2021multiagent,brittain2021autonomous}. An action that resolves one encounter can therefore create a new conflict, intensify another encounter, or transfer the conflict to a neighboring aircraft. Under degraded surveillance conditions, these interactions become more difficult to manage because each aircraft must coordinate its response using information that is uncertain, delayed, incomplete, or temporarily unavailable. Effective local decision-making must therefore account for both the current conflict and the effects of each maneuver on multiple surrounding aircraft.

\subsection{Study Objective and Contributions}

To address these challenges, this study develops a Deep Q-Network (DQN)-based Multi-Agent Reinforcement Learning (MARL) model for decentralized conflict resolution in a degraded surveillance environment. Small UAVs and eVTOL aircraft are represented as autonomous agents in a structured three-dimensional simulation environment with separate altitude layers, directional traffic flows, cruise lanes, and passing lanes. The environment generates traffic scenarios across different aircraft densities and minimum-separation thresholds. Each agent observes its own operating state and the available information about nearby aircraft and selects a discrete maneuvering action. Degraded surveillance is represented through observation noise, communication delay, information dropout, and unavailable neighbor information. The model also incorporates wind disturbance, actuator uncertainty, model uncertainty, aircraft-specific motion constraints, corridor boundaries, minimum-separation requirements, and maneuver-dependent energy consumption. Separate DQN policies are trained for UAV and eVTOL agents to account for differences in speed, altitude layer, maneuvering limits, and energy characteristics.

The agents are trained to maintain safe separation, avoid collisions, remain within the corridor, follow the assigned traffic direction, and reduce unnecessary maneuvers. The trained policies are evaluated across the generated traffic scenarios to examine safety performance and the use of maintain, speed-control, turning, and vertical actions in safe and loss-of-separation (LOS) states. The model provides an aircraft-level conflict-resolution capability when surveillance information is inaccurate, delayed, incomplete, or unavailable. The results help assess how aircraft density, separation requirements, and degraded observations influence conflict-resolution performance in AAM operations.

\section{Literature Review}
\label{sec:marl_literature_review}

Aircraft conflict detection and resolution has traditionally been addressed using geometric methods, rule-based logic, velocity-obstacle techniques, optimization models, force-field methods, and Markov decision process formulations \cite{kuchar2000review,fiorini1998velocity,kochenderfer2012acasx}. These approaches established the foundation for separation assurance by identifying potential conflicts and determining suitable avoidance maneuvers. However, many of them rely on accurate traffic information, predefined encounter rules, or centralized coordination. Their application becomes more difficult in dense multi-aircraft environments because a maneuver selected by one aircraft immediately changes the relative motion and available responses of surrounding aircraft.

To reduce conflicts before they occur, several studies have focused on strategic trajectory planning. Dai et al. \cite{dai2021conflictfree} developed a four-dimensional path-planning approach that considered both static and dynamic airspace conflicts, while Tang et al. \cite{tang2021automated} proposed an automated flight-planning framework for high-density UAM operations. These methods can reduce planned conflicts and improve traffic organization. However, tactical onboard conflict resolution is still required when aircraft deviate from planned trajectories or when uncertain operating conditions change the encounter after departure.

Reinforcement learning offers a different approach by allowing an aircraft to learn a maneuver-selection policy through repeated interaction with a simulated environment. The Deep Q-Network introduced by Mnih et al. \cite{mnih2015dqn} extended Q-learning to high-dimensional observations and discrete action spaces through neural networks, experience replay, and target networks. The MARL extends this process to environments in which several agents make decisions simultaneously. This is relevant to aircraft conflict resolution because the action of one aircraft affects the observations, risks, and available actions of nearby aircraft. At the same time, MARL introduces nonstationarity because the environment observed by one agent changes as the policies of the other agents evolve \cite{lowe2017multiagent}.

Building on these developments, several studies have applied MARL to aircraft separation and conflict management. Brittain et al. \cite{brittain2021autonomous} developed a deep MARL model for autonomous separation assurance in high-density stochastic airspace and showed that decentralized policies can support conflict resolution without continuous tactical intervention from a centralized controller. Their later distributed framework addressed encounters involving a variable number of aircraft rather than a fixed pairwise structure \cite{brittain2021onetoany}. Isufaj et al. \cite{isufaj2022multi} used graph convolutional reinforcement learning for multi-UAV conflict resolution and demonstrated that multi-aircraft conflicts cannot always be decomposed into independent pairwise encounters. Huang et al. \cite{huang2023strategic} developed a recurrent MARL model for strategic conflict management in UAM operations under uncertainty, while Deniz and Wang \cite{deniz2024autonomous} applied deep MARL to autonomous conflict resolution among eVTOL aircraft in three-dimensional airspace. A related study by Deniz and Wang \cite{deniz2024coordination} examined vehicle coordination in structured high-density AAM operations. Together, these studies demonstrate the potential of MARL for environments in which several aircraft make interdependent decisions.

Other learning-based studies have examined decentralized control using locally available observations. Waltz et al. \cite{waltz2024selforganized} developed a self-organized eVTOL arrival system in which each aircraft followed a shared policy and selected actions using local information. Their simulations also examined robustness to sensor noise and changing inbound traffic patterns. This work supports decentralized decision-making when complete global traffic information is unavailable, although its focus on arrival management differs from en-route conflict resolution within a structured corridor.

In addition to the learning method, the design of the operating airspace directly affects conflict frequency and the complexity of resolution. Structured airspace concepts use corridors, routes, altitude layers, and directional traffic flows to organize low-altitude operations. Thipphavong et al. \cite{thipphavong2018uam} identified structured procedures and traffic-management concepts as important components of UAM operations. Bauranov and Rakas \cite{bauranov2021airspace} reviewed corridor-, layer-, sector-, and route-based airspace designs, while Doole et al. \cite{doole2021constrained} examined vertical and horizontal traffic organization for large-scale drone operations.

The effect of structure becomes more important as traffic demand increases. Wang et al. \cite{wang2021trafficassignment} showed that intensive UAM operations require traffic-assignment methods that account for network organization and congestion. Their findings indicate that airspace design influences where traffic demand is concentrated and where conflicts are likely to occur. Similarly, Taye et al. \cite{taye2024safe} proposed a real-time trajectory-planning framework that updates aircraft trajectories as traffic conditions change. These studies support the use of structured corridors for evaluating lane compliance, traffic interactions, and tactical conflict-resolution actions.

Even within an organized airspace, the effectiveness of conflict resolution depends on the quality of surveillance information available to the aircraft. Sinopoli et al. \cite{sinopoli2004kalman} showed how intermittent observations affect state estimation over unreliable communication channels, while Hespanha et al. \cite{hespanha2007networked} examined the effects of communication delay and packet loss in networked control systems. Reinforcement-learning studies under perturbed or delayed observations have also shown that measurement errors and stale information can lead to poor action selection \cite{zhang2020robustdrl,ramstedt2021randomdelays,agarwal2021delayedobservations}.

This issue is directly relevant to AAM because conflict detection depends on both tracking quality and encounter geometry. Dai et al. \cite{dai2024tracking} developed a probabilistic framework to evaluate how tracking uncertainty and airspace design affect tactical conflict-detection effectiveness. Their findings show that surveillance performance cannot be evaluated independently of traffic configuration and airspace organization. This supports representing observation noise, communication delay, and information dropout directly within the conflict-resolution environment rather than assuming continuously accurate and current aircraft states.

Aircraft performance and energy availability further constrain the actions that can be used to resolve a conflict. Turning, acceleration, vertical movement, and landing require different levels of maneuvering capability and energy. Gong et al. \cite{gong2023power} developed maneuver-dependent power models for multirotor aircraft, while Dai et al. \cite{dai2022dataefficient} examined power estimation for quadrotor and eVTOL operations. Pradeep and Wei \cite{pradeep2019energyefficient} also showed that multirotor eVTOL trajectories must balance timing requirements with energy consumption. These findings support including remaining energy in the aircraft state so that the learning policy can consider both immediate safety and the aircraft's remaining operational capability.

Although the existing literature provides a strong foundation for MARL-based aircraft conflict resolution, important gaps remain. Many studies focus on one aircraft category, one type of operating environment, or sufficiently accurate and current traffic information. Strategic path-planning and traffic-assignment studies often address conflicts before departure or through centralized coordination, while tactical MARL studies commonly assume homogeneous aircraft or simplified observation conditions. Structured corridor operation, heterogeneous UAV and eVTOL traffic, observation noise, communication delay, information dropout, aircraft dynamics, energy consumption, lane containment, and minimum-separation requirements are not commonly represented within a single framework. In addition, learned policies are often evaluated under a limited range of traffic densities and separation conditions.

This study addresses these gaps by developing a DQN-based MARL model for heterogeneous UAV and eVTOL operations under degraded surveillance conditions. The model integrates a structured three-dimensional corridor, aircraft-specific operating characteristics, uncertain and delayed observations, motion and energy constraints, and multiple conflict-resolution actions. The trained policies are evaluated across different traffic densities and minimum-separation thresholds to examine safety performance, LOS frequency and duration, speed behavior, and aircraft action-selection patterns.

\section{Methodology}
\label{sec:marl_methodology}

This section presents the methodology used to develop, train, and evaluate the MARL-based aircraft conflict-resolution model for AAM traffic in air corridors under degraded surveillance. The methodology consists of three main parts. First, the simulation environment is developed by defining the structured airspace corridor, heterogeneous traffic-generation process, aircraft dynamics, uncertainty models, and energy-consumption model. Second, the MARL decision-making model is formulated by defining the state space, action space, safety metrics, reward function, and DQN architecture. Third, the training and testing procedures are described to explain how the UAV and eVTOL policies are learned and evaluated. 

\begin{longtable}{p{0.22\textwidth}p{0.72\textwidth}}
\caption{Notation used in the MARL model.}
\label{tab:marl_notation}\\
\hline
\textbf{Notation} & \textbf{Description} \\
\hline
\endfirsthead

\hline
\textbf{Notation} & \textbf{Description} \\
\hline
\endhead

\hline
\endfoot

\multicolumn{2}{l}{\textbf{Airspace corridor geometry}} \\
\hline
$P_1$ & Entry waypoint of an operational lane. \\
$P_2$ & Exit waypoint of an operational lane. \\
$(x,y,z)$ & Three-dimensional Cartesian coordinates of an aircraft or waypoint. \\
$L_c$ & Length of the operational lane or corridor. \\
$W_c$ & Width of the operational lane. \\
$H_c$ & Height of the operational lane. \\
$\mathbf{d}$ & Longitudinal unit direction vector along the lane centerline. \\
$\mathbf{u}$ & Global vertical unit vector. \\
$\mathbf{s}$ & Lateral unit vector of the lane coordinate system. \\
$\mathbf{r}$ & Relative position vector from the lane entry waypoint to the aircraft position. \\
$\ell$ & Longitudinal coordinate of an aircraft in the local lane coordinate system. \\
$w$ & Lateral offset of an aircraft from the lane centerline. \\
$h$ & Vertical offset of an aircraft from the lane reference altitude. \\

\hline
\multicolumn{2}{l}{\textbf{Traffic generation and initialization}} \\
\hline
$N_{\mathrm{eVTOL}}$ & Number of eVTOL aircraft in an episode. \\
$N_{\mathrm{UAV}}$ & Number of UAVs in an episode. \\
$N$ & Total number of aircraft in an episode. \\
$\zeta_i$ & Aircraft type of agent $i$. \\
$\mathcal{U}_{\mathbb{Z}}(a,b)$ & Discrete uniform distribution over integer values from $a$ to $b$. \\
$g$ & Aircraft group associated with one aircraft type and one operating lane. \\
$N_g$ & Number of aircraft in group $g$. \\
$L_{\mathrm{use}}$ & Usable corridor length for aircraft initialization. \\
$L_{\mathrm{buf}}$ & Buffer distance removed from the corridor length during initialization. \\
$\Delta_g$ & Nominal aircraft spacing for group $g$. \\
$\bar{\Delta}_g$ & Enforced aircraft spacing after applying the minimum spacing rule. \\
$\Delta_{\min}$ & Minimum spacing used during initial aircraft placement. \\
$k$ & Aircraft index within group $g$. \\
$\ell_{k,g}$ & Longitudinal initialization distance of the $k$th aircraft in group $g$. \\
$\ell_0$ & Initial offset from the lane starting point. \\
$\epsilon_{k,g}$ & Random longitudinal perturbation used during aircraft initialization. \\
$P_i(0)$ & Initial position of aircraft $i$. \\
$P_g^{\mathrm{start}}$ & Starting waypoint of the assigned operating lane for group $g$. \\
$\mathbf{d}_g$ & Travel direction vector of group $g$. \\
$V_i^0$ & Nominal speed of aircraft $i$. \\
$V_{\mathrm{eVTOL}}$ & Nominal speed of an eVTOL aircraft. \\
$V_{\mathrm{UAV}}$ & Nominal speed of a UAV. \\
$\eta_i$ & Random speed variation factor for aircraft $i$. \\
$\mathbf{v}_i(0)$ & Initial velocity vector of aircraft $i$. \\
$E_i(0)$ & Initial energy of aircraft $i$. \\
$E_i^{\max}$ & Maximum available energy of aircraft $i$. \\
$\rho_i$ & Initial energy ratio assigned to aircraft $i$. \\

\hline
\multicolumn{2}{l}{\textbf{Aircraft dynamics and maneuvers}} \\
\hline
$\mathbf{p}_{i,t}$ & Position vector of aircraft $i$ at time step $t$. \\
$\mathbf{v}_{i,t}$ & Velocity vector of aircraft $i$ at time step $t$. \\
$\Delta t$ & Simulation time step. \\
$\mathbf{v}_{i,t}^{a}$ & Velocity vector of aircraft $i$ after action execution. \\
$V_{i,t}$ & Speed of aircraft $i$ at time step $t$. \\
$\omega_i$ & Turn rate of aircraft $i$. \\
$g$ & Gravitational acceleration. \\
$\phi_i$ & Maximum bank angle of aircraft $i$. \\
$\phi_{\mathrm{eVTOL}}$ & Maximum bank angle for eVTOL aircraft. \\
$\phi_{\mathrm{UAV}}$ & Maximum bank angle for UAVs. \\
$\theta_i$ & Heading-change angle over one time step for aircraft $i$. \\
$\mathbf{R}(\theta_i)$ & Horizontal rotation matrix for turning maneuvers. \\
$\psi_{i,t}$ & Heading angle of aircraft $i$ at time step $t$. \\
$\gamma_{i,t}$ & Flight-path angle of aircraft $i$ at time step $t$. \\
$\gamma_{i,t}^{a}$ & Updated flight-path angle after a vertical maneuver. \\
$\delta_{\gamma}$ & Flight-path angle increment used for climb or descent. \\
$\gamma_i^{\max}$ & Maximum flight-path angle of aircraft $i$. \\
$\gamma_{\mathrm{eVTOL}}^{\max}$ & Maximum flight-path angle for eVTOL aircraft. \\
$\gamma_{\mathrm{UAV}}^{\max}$ & Maximum flight-path angle for UAVs. \\

\hline
\multicolumn{2}{l}{\textbf{Uncertainty modeling}} \\
\hline
$\tilde{\mathbf{v}}_{i,t}$ & Perturbed velocity vector after process uncertainty is applied. \\
$\mathbf{w}_{i,t}$ & Wind disturbance vector for aircraft $i$. \\
$\boldsymbol{\eta}_{i,t}^{a}$ & Actuator uncertainty vector. \\
$\boldsymbol{\eta}_{i,t}^{m}$ & Model uncertainty vector. \\
$\sigma_w$ & Standard deviation of wind disturbance. \\
$\sigma_a$ & Standard deviation of actuator uncertainty. \\
$\sigma_m$ & Standard deviation of model uncertainty. \\
$\mathbf{I}$ & Identity matrix. \\
$\odot$ & Element-wise multiplication operator. \\
$\mathbf{v}_{i,t}^{b}$ & Velocity vector after multiplicative actuator uncertainty. \\
$c_d$ & Drag coefficient. \\
$\hat{\mathbf{p}}_{i,t}$ & Observed position of aircraft $i$. \\
$\boldsymbol{\epsilon}^{p}_{i,t}$ & Position measurement error for aircraft $i$. \\
$\sigma_{p,\mathrm{self}}$ & Standard deviation of self-position measurement error. \\
$\hat{V}_{i,t}$ & Observed speed of aircraft $i$. \\
$\hat{\psi}_{i,t}$ & Observed heading angle of aircraft $i$. \\
$\hat{\gamma}_{i,t}$ & Observed flight-path angle of aircraft $i$. \\
$\epsilon^V_{i,t}$ & Speed measurement error. \\
$\epsilon^{\psi}_{i,t}$ & Heading-angle measurement error. \\
$\epsilon^{\gamma}_{i,t}$ & Flight-path-angle measurement error. \\
$\hat{\mathbf{v}}_{i,t}$ & Observed velocity vector of aircraft $i$. \\
$\hat{E}_{i,t}$ & Observed energy level of aircraft $i$. \\
$\epsilon^E_{i,t}$ & Relative energy measurement error. \\
$\tau_{ij,t}$ & Communication delay for observing aircraft $j$ by aircraft $i$. \\
$\tau_{\max}$ & Maximum communication delay. \\
$\mathbf{p}_{j,t}^{\mathrm{delay}}$ & Delayed position estimate of neighboring aircraft $j$. \\
$P_{\mathrm{drop}}$ & Probability of information dropout. \\

\hline
\multicolumn{2}{l}{\textbf{Energy-consumption model}} \\
\hline
$E_{i,t}$ & Remaining energy of aircraft $i$ at time step $t$. \\
$P_{i,t}$ & Power consumption of aircraft $i$ at time step $t$. \\
$P_{\mathrm{forward}}$ & Power consumption during forward flight. \\
$P_{\mathrm{hover}}$ & Hover power. \\
$P_{\mathrm{blade}}$ & Blade profile power component. \\
$P_{\mathrm{induced}}$ & Induced power component. \\
$P_{\mathrm{vertical}}^{\mathrm{up}}$ & Power consumption during climbing flight. \\
$P_{\mathrm{vertical}}^{\mathrm{down}}$ & Power consumption during descending flight. \\
$P_{\mathrm{turn}}$ & Power consumption during turning flight. \\
$P_1,P_2,P_3,P_4$ & Power components used in the forward-flight power model. \\
$P_{1,r},P_{2,r},P_{3,r},P_{4,r}$ & Rotor-specific power components during turning flight. \\
$W$ & Aircraft weight. \\
$n$ & Number of rotors. \\
$\rho$ & Air density. \\
$A$ & Rotor disk area. \\
$\delta$ & Profile drag coefficient. \\
$s$ & Rotor solidity. \\
$C_T$ & Thrust coefficient. \\
$W_r$ & Rotor load. \\
$k$ & Induced power correction factor. \\
$v_0$ & Induced velocity in hover. \\
$S_{\mathrm{fp,par}}$ & Equivalent flat plate area in forward flight. \\
$S_{\mathrm{fp,perp}}$ & Equivalent flat plate area in vertical flight. \\

\hline
\multicolumn{2}{l}{\textbf{MARL state and action space}} \\
\hline
$\mathbf{s}_{i,t}^{\mathrm{self}}$ & Self-state vector of aircraft $i$ at time step $t$. \\
$\mathbf{s}_{ij,t}^{\mathrm{nbr}}$ & Neighbor-state vector of aircraft $j$ observed by aircraft $i$. \\
$\Delta x_{ij,t}$ & Relative $x$-position of aircraft $j$ with respect to aircraft $i$. \\
$\Delta y_{ij,t}$ & Relative $y$-position of aircraft $j$ with respect to aircraft $i$. \\
$\Delta z_{ij,t}$ & Relative $z$-position of aircraft $j$ with respect to aircraft $i$. \\
$\mathcal{N}_{i,t}$ & Set of neighboring aircraft observed by aircraft $i$ at time step $t$. \\
$R_{\mathrm{local}}$ & Local observation radius. \\
$K$ & Maximum number of nearest neighbors included in the state vector. \\
$\mathbf{S}_{i,t}$ & Complete state vector of aircraft $i$ at time step $t$. \\
$d_s$ & Dimension of the state vector. \\
$\mathcal{A}$ & Discrete action space. \\
$a_{i,t}$ & Action selected by aircraft $i$ at time step $t$. \\

\hline
\multicolumn{2}{l}{\textbf{Safety metrics and reward function}} \\
\hline
$d_{ij,t}$ & Relative distance between aircraft $i$ and aircraft $j$ at time step $t$. \\
$D_{\zeta_i}^{\min}$ & Minimum-separation threshold associated with aircraft type $\zeta_i$. \\
$D_{\mathrm{eVTOL}}^{\min}$ & Minimum-separation threshold for eVTOL aircraft. \\
$D_{\mathrm{UAV}}^{\min}$ & Minimum-separation threshold for UAVs. \\
$L_{ij,t}$ & Pairwise loss-of-separation indicator for aircraft pair $(i,j)$. \\
$L_{i,t}$ & Number of loss-of-separation violations involving aircraft $i$. \\
$D^{\mathrm{col}}$ & Collision-distance threshold. \\
$C_{ij,t}$ & Pairwise collision indicator for aircraft pair $(i,j)$. \\
$C_{i,t}$ & Number of collision events involving aircraft $i$. \\
$R_{i,t}$ & Total reward received by aircraft $i$ at time step $t$. \\
$R_{i,t}^{\mathrm{lane}}$ & Lane-containment reward. \\
$R_{i,t}^{\mathrm{lat}}$ & Lateral-deviation reward term. \\
$P_{i,t}^{\mathrm{LOS}}$ & Loss-of-separation penalty. \\
$P_{i,t}^{\mathrm{col}}$ & Collision penalty. \\
$I_{i,t}^{\mathrm{safe}}$ & Indicator showing whether aircraft $i$ has no LOS or collision. \\
$\hat{\mathbf{v}}_{i,t}$ & Unit velocity vector of aircraft $i$. \\
$\mathbf{d}_i$ & Desired flow direction of the assigned lane for aircraft $i$. \\
$\alpha_{i,t}$ & Alignment value between aircraft velocity and desired lane direction. \\
$e_{i,t}^{\mathrm{lat}}$ & Lateral deviation from the desired flow direction. \\
$I_{i,t}^{\mathrm{lane}}$ & Indicator showing whether aircraft $i$ is inside its assigned lane. \\

\hline
\multicolumn{2}{l}{\textbf{Deep Q-Network Architecture}} \\
\hline
$Q(\mathbf{S}_{i,t},a_{i,t};\theta)$ & Q-value function approximated by the DQN. \\
$\mathbf{S}_{i,t}$ & State vector of aircraft $i$ at time step $t$. \\
$a_{i,t}$ & Action selected by aircraft $i$ at time step $t$. \\
$\theta$ & Policy-network parameters. \\
$\mathbf{q}_{i,t}$ & Vector of Q-values produced by the DQN for aircraft $i$. \\
$f_{\theta}(\cdot)$ & Neural network mapping from the state vector to the Q-value vector. \\
$Q(\mathbf{S}_{i,t},0;\theta)$ & Q-value for action 0 in state $\mathbf{S}_{i,t}$. \\
$Q(\mathbf{S}_{i,t},1;\theta)$ & Q-value for action 1 in state $\mathbf{S}_{i,t}$. \\
$Q(\mathbf{S}_{i,t},5;\theta)$ & Q-value for action 5 in state $\mathbf{S}_{i,t}$. \\
$a_{i,t}^{*}$ & Greedy action selected using the maximum Q-value. \\

\hline
\multicolumn{2}{l}{\textbf{Training Procedure}} \\
\hline
$N_{\mathrm{ep}}$ & Total number of training episodes. \\
$e$ & Training episode index, where $e=1,2,\ldots,N_{\mathrm{ep}}$. \\
$T$ & Number of simulation time steps per episode. \\
$i$ & Aircraft agent index. \\
$t$ & Simulation time-step index. \\
$\epsilon$ & Exploration probability in the $\epsilon$-greedy policy. \\
$\epsilon_e$ & Exploration probability in training episode $e$. \\
$\epsilon_0$ & Initial exploration probability. \\
$\epsilon_{\min}$ & Minimum exploration probability. \\
$\beta$ & Exploration decay factor. \\
$\mathbf{S}_{i,t}$ & Current state vector of aircraft $i$ at time step $t$. \\
$a_{i,t}$ & Action selected by aircraft $i$ at time step $t$. \\
$r_{i,t}$ & Reward received by aircraft $i$ at time step $t$. \\
$\mathbf{S}_{i,t+1}$ & Next state vector of aircraft $i$ after executing action $a_{i,t}$. \\
$d_{i,t}$ & Terminal indicator for aircraft $i$ at time step $t$. \\
$\left(\mathbf{S}_{i,t},a_{i,t},r_{i,t},\mathbf{S}_{i,t+1},d_{i,t}\right)$ & Transition stored in the replay buffer. \\
$C_{\mathrm{buffer}}$ & Maximum capacity of the replay buffer. \\
$B$ & Mini-batch size sampled from the replay buffer. \\
$y_{i,t}$ & Target Q-value used in the DQN update. \\
$a'$ & Candidate next action used in the Bellman target. \\
$\theta^{-}$ & Target-network parameters. \\
$Q(\mathbf{S}_{i,t+1},a';\theta^{-})$ & Target-network Q-value for the next state and candidate action. \\
$\gamma$ & Discount factor used to weight future rewards. \\
$\mathcal{L}(\theta)$ & DQN loss function minimized during training. \\
$\alpha_{\mathrm{lr}}$ & Learning rate used by the Adam optimizer. \\
$N_{\mathrm{target}}$ & Target-network update interval. \\

\hline
\multicolumn{2}{l}{\textbf{Testing scenario design}} \\
\hline
$\mathcal{D}$ & Set of aircraft-density levels used in testing. \\
$\mathcal{S}$ & Set of minimum-separation thresholds used in testing. \\
$N_{\mathrm{test}}$ & Total number of base testing scenarios. \\
$d$ & Aircraft-density level in a testing scenario, where $d \in \mathcal{D}$. \\
$s$ & Minimum-separation threshold in a testing scenario, where $s \in \mathcal{S}$. \\
$m$ & Repeated test episode index for a given base testing scenario. \\
$\mathbf{p}_{i,0}^{(m)}$ & Initial position of aircraft $i$ in repeated test episode $m$. \\
$x_{i,0}^{(m)}$ & Initial longitudinal position of aircraft $i$ in repeated test episode $m$. \\
$y_{i,0}^{(m)}$ & Initial lateral position of aircraft $i$ in repeated test episode $m$. \\
$z_{i,0}^{(m)}$ & Initial altitude of aircraft $i$ in repeated test episode $m$. \\
\hline

\end{longtable}

\subsection{Simulation Environment Design}
\label{subsec:simulation_environment_design}

The simulation environment represents a structured AAM corridor system in which heterogeneous UAV and eVTOL traffic operates under degraded surveillance conditions. The environment defines the physical operating space, aircraft initialization process, aircraft motion, uncertainty effects, and energy-consumption behavior. These components provide the basis for generating training and testing episodes used by the MARL agents.

\subsubsection{Airspace Corridor Design}
\label{par:airspace_corridor_design}

\begin{figure}[h]
\centering
\includegraphics[width=14cm,height=12cm]{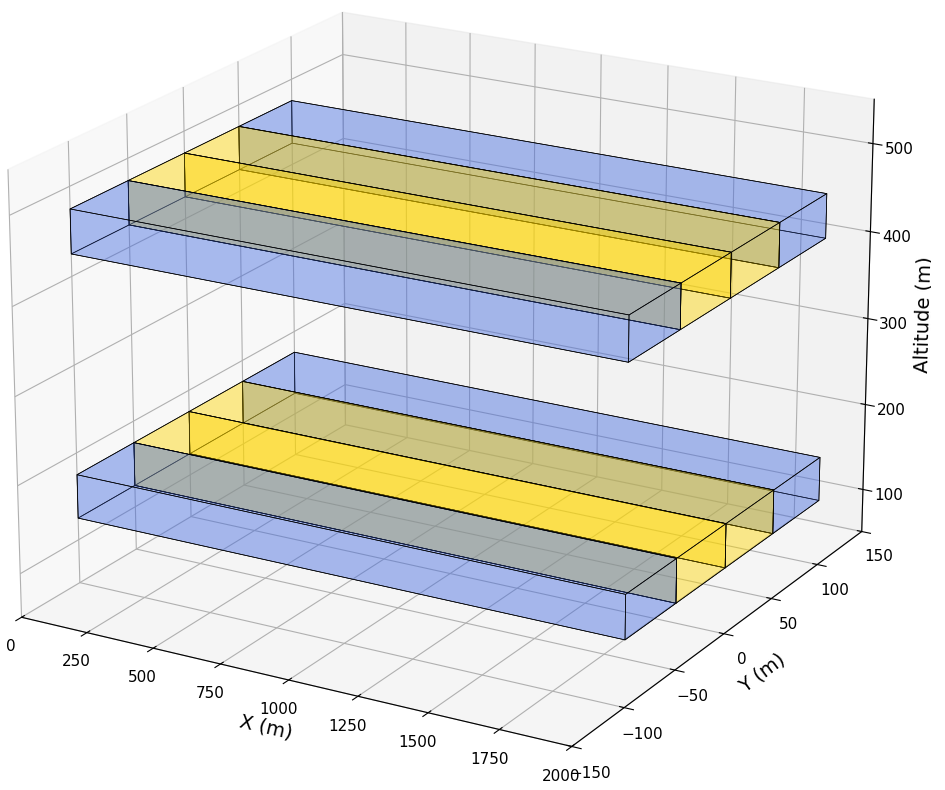}
\caption{Structured air corridor used in the MARL environment. 
}
\label{fig:corridor_architecture}
\end{figure}

A structured three-dimensional corridor was developed as the operating environment for the MARL model. The airspace contains two vertically separated layers: a UAV layer centered at 150~m and an eVTOL layer centered at 450~m, with 300~m vertical separation. Each layer contains four parallel lanes: two cruise lanes and two passing lanes. Lane~A supports traffic from the corridor entry to the exit, while Lane~B supports traffic in the opposite direction. Each lane is 2,000~m long, 50~m wide, and 50~m high, resulting in eight operational lanes across the two layers. Figure~\ref{fig:corridor_architecture} shows the corridor configuration.

Each lane is modeled as a three-dimensional rectangular prism. Let $P_1=(x_1,y_1,z_1)$ and $P_2=(x_2,y_2,z_2)$ denote the lane entry and exit waypoints. The lane length and longitudinal unit vector are

\begin{equation}
L_c=\left|P_2-P_1\right|,
\qquad
\mathbf{d}=\frac{P_2-P_1}{\left|P_2-P_1\right|}.
\label{eq:lane_length_direction}
\end{equation}

The vertical and lateral unit vectors are defined as

\begin{equation}
\mathbf{u}=
\begin{bmatrix}
0 & 0 & 1
\end{bmatrix}^{T},
\qquad
\mathbf{s}=
\frac{\mathbf{d}\times\mathbf{u}}
{\left|\mathbf{d}\times\mathbf{u}\right|}.
\label{eq:lane_unit_vectors}
\end{equation}

For an aircraft at position $P=(x,y,z)$, the relative position vector is $\mathbf{r}=P-P_1$. Its local lane coordinates are

\begin{equation}
\ell=\mathbf{r}\cdot\mathbf{d},
\qquad
w=\mathbf{r}\cdot\mathbf{s},
\qquad
h=\mathbf{r}\cdot\mathbf{u},
\label{eq:local_lane_coordinates}
\end{equation}

where $\ell$, $w$, and $h$ represent the longitudinal, lateral, and vertical coordinates, respectively. With $L_c=2000~\mathrm{m}$, $W_c=50~\mathrm{m}$, and $H_c=50~\mathrm{m}$, an aircraft remains inside its assigned lane when

\begin{equation}
0\leq\ell\leq L_c,
\qquad
-\frac{W_c}{2}\leq w\leq\frac{W_c}{2},
\qquad
-\frac{H_c}{2}\leq h\leq\frac{H_c}{2}.
\label{eq:lane_containment}
\end{equation}

These conditions are evaluated at each simulation step to identify whether an aircraft remains within its assigned lane.

\subsubsection{Heterogeneous Traffic Generation}
\label{par:heterogeneous_traffic_generation}

Each training episode includes a heterogeneous mix of UAVs and eVTOL aircraft with randomized counts, positions, speeds, and energy levels. The number of aircraft of each type is sampled independently as

\begin{equation}
N_{\mathrm{eVTOL}} \sim \mathcal{U}_{\mathbb{Z}}(4,18),
\qquad
N_{\mathrm{UAV}} \sim \mathcal{U}_{\mathbb{Z}}(4,18),
\label{eq:aircraft_counts}
\end{equation}

where $\mathcal{U}_{\mathbb{Z}}(a,b)$ denotes a discrete uniform distribution over the integers from $a$ to $b$. The total number of aircraft is

\begin{equation}
N = N_{\mathrm{eVTOL}} + N_{\mathrm{UAV}}.
\label{eq:total_aircraft}
\end{equation}

The aircraft type of agent $i$ is defined as

\begin{equation}
\zeta_i =
\begin{cases}
\mathrm{eVTOL}, & 1 \leq i \leq N_{\mathrm{eVTOL}},\\
\mathrm{UAV}, & N_{\mathrm{eVTOL}} < i \leq N.
\end{cases}
\label{eq:aircraft_type_assignment}
\end{equation}

Aircraft are initialized along the centerline of their assigned lanes. For aircraft group $g$, defined by aircraft type and lane assignment, the usable corridor length and nominal spacing are

\begin{equation}
L_{\mathrm{use}} = L_c - L_{\mathrm{buf}},
\qquad
\Delta_g = \frac{L_{\mathrm{use}}}{N_g+1},
\label{eq:usable_length_spacing}
\end{equation}

where $L_c=2000~\mathrm{m}$ and $L_{\mathrm{buf}}=300~\mathrm{m}$. The spacing used for initialization is

\begin{equation}
\bar{\Delta}_g
=
\max\left(\Delta_g,\Delta_{\min}\right),
\qquad
\Delta_{\min}=300~\mathrm{m}.
\label{eq:minimum_spacing_rule}
\end{equation}

The longitudinal position of the $k$th aircraft in group $g$ is

\begin{equation}
\ell_{k,g}
=
\ell_0+(k+1)\bar{\Delta}_g+\epsilon_{k,g},
\qquad
\epsilon_{k,g}\sim\mathcal{U}(-20,20),
\label{eq:longitudinal_initial_distance}
\end{equation}

where $\ell_0=100~\mathrm{m}$. The initial position is

\begin{equation}
P_i(0)
=
P_g^{\mathrm{start}}
+
\ell_{k,g}\mathbf{d}_g,
\label{eq:initial_position}
\end{equation}

where $P_g^{\mathrm{start}}$ and $\mathbf{d}_g$ denote the starting waypoint and travel-direction vector of the assigned lane.

The nominal speed depends on aircraft type:

\begin{equation}
V_i^0 =
\begin{cases}
26~\mathrm{m/s}, & \zeta_i=\mathrm{eVTOL},\\
7~\mathrm{m/s}, & \zeta_i=\mathrm{UAV}.
\end{cases}
\label{eq:nominal_speed}
\end{equation}

The initial speed and velocity are

\begin{equation}
V_i(0)=V_i^0\eta_i,
\qquad
\eta_i\sim\mathcal{U}(0.90,1.10),
\label{eq:initial_speed}
\end{equation}

\begin{equation}
\mathbf{v}_i(0)=V_i(0)\mathbf{d}_g.
\label{eq:initial_velocity}
\end{equation}

The initial energy level is sampled as

\begin{equation}
E_i(0)=E_i^{\max}\rho_i,
\qquad
\rho_i\sim\mathcal{U}(0.60,1.00),
\label{eq:initial_energy}
\end{equation}

where

\begin{equation}
E_i^{\max}
=
\begin{cases}
E_{\mathrm{eVTOL}}^{\max}, & \zeta_i=\mathrm{eVTOL},\\
E_{\mathrm{UAV}}^{\max}, & \zeta_i=\mathrm{UAV}.
\end{cases}
\label{eq:max_energy}
\end{equation}

This procedure generates different traffic densities and initial operating conditions across training episodes.

\subsubsection{Aircraft Dynamics, Uncertainty, and Energy Modeling}
\label{sec:aircraft_dynamics_uncertainty_energy}

This section describes the aircraft motion model, uncertainty representation, and energy-consumption model used in the MARL environment. The models are applied to both UAV and eVTOL agents during each simulation time step.

\vspace{0.2cm}

\noindent \textbf{Aircraft Dynamic Model}
\label{subsec:aircraft_dynamic_model}

Each aircraft is represented by its position and velocity vectors,

\begin{equation}
\mathbf{p}_{i,t}
=
\begin{bmatrix}
x_{i,t} & y_{i,t} & z_{i,t}
\end{bmatrix}^{T},
\qquad
\mathbf{v}_{i,t}
=
\begin{bmatrix}
v_{x,i,t} & v_{y,i,t} & v_{z,i,t}
\end{bmatrix}^{T}.
\label{eq:aircraft_state_vectors}
\end{equation}

With a simulation time step of $\Delta t=1~\mathrm{s}$, the position is updated as

\begin{equation}
\mathbf{p}_{i,t+1}
=
\mathbf{p}_{i,t}
+
\mathbf{v}_{i,t+1}\Delta t.
\label{eq:position_update}
\end{equation}

The velocity is modified according to the selected maneuver. For a turning action, the aircraft speed and turn rate are

\begin{equation}
V_{i,t}=\left\|\mathbf{v}_{i,t}\right\|,
\qquad
\omega_i=\frac{g\tan(\phi_i)}{V_{i,t}},
\label{eq:speed_turn_rate}
\end{equation}

where

\begin{equation}
\phi_i=
\begin{cases}
15^{\circ}, & \zeta_i=\mathrm{eVTOL},\\
20^{\circ}, & \zeta_i=\mathrm{UAV}.
\end{cases}
\label{eq:bank_angle_type}
\end{equation}

The heading change is $\theta_i=\omega_i\Delta t$, and the post-turn velocity is

\begin{equation}
\mathbf{v}_{i,t}^{a}
=
\mathbf{R}(\pm\theta_i)\mathbf{v}_{i,t},
\qquad
\mathbf{R}(\theta_i)
=
\begin{bmatrix}
\cos\theta_i & -\sin\theta_i & 0\\
\sin\theta_i & \cos\theta_i & 0\\
0 & 0 & 1
\end{bmatrix},
\label{eq:turn_velocity}
\end{equation}

where the positive and negative signs represent left and right turns, respectively.

For vertical maneuvers, the heading and flight-path angles are

\begin{equation}
\psi_{i,t}
=
\tan^{-1}\left(\frac{v_{y,i,t}}{v_{x,i,t}}\right),
\qquad
\gamma_{i,t}
=
\tan^{-1}\left(
\frac{v_{z,i,t}}
{\sqrt{v_{x,i,t}^{2}+v_{y,i,t}^{2}}}
\right).
\label{eq:heading_flight_path_angles}
\end{equation}

The updated flight-path angle is

\begin{equation}
\gamma_{i,t}^{a}
=
\operatorname{clip}
\left(
\gamma_{i,t}+\delta_{\gamma},
-\gamma_i^{\max},
\gamma_i^{\max}
\right),
\qquad
\delta_{\gamma}=\pm3^{\circ},
\label{eq:updated_gamma}
\end{equation}

where

\begin{equation}
\gamma_i^{\max}
=
\begin{cases}
15^{\circ}, & \zeta_i=\mathrm{eVTOL},\\
10^{\circ}, & \zeta_i=\mathrm{UAV}.
\end{cases}
\label{eq:gamma_type}
\end{equation}

The resulting velocity is

\begin{equation}
\mathbf{v}_{i,t}^{a}
=
V_{i,t}
\begin{bmatrix}
\cos(\gamma_{i,t}^{a})\cos(\psi_{i,t})\\
\cos(\gamma_{i,t}^{a})\sin(\psi_{i,t})\\
\sin(\gamma_{i,t}^{a})
\end{bmatrix}.
\label{eq:vertical_maneuver_velocity}
\end{equation}

\noindent \textbf{Uncertainty Modeling}
\label{subsec:uncertainty_modeling}

Process uncertainty is represented through wind, actuator, model, and drag effects. The updated velocity is

\begin{equation}
\mathbf{v}_{i,t+1}
=
(1-c_d)
\left[
\left(
\mathbf{v}_{i,t}^{a}
+
\mathbf{w}_{i,t}
\right)
\odot
\left(
\mathbf{1}+\boldsymbol{\eta}_{i,t}^{a}
\right)
+
\boldsymbol{\eta}_{i,t}^{m}
\right],
\label{eq:uncertain_velocity_update}
\end{equation}

where $c_d=0.01$ and

\begin{equation}
\mathbf{w}_{i,t}\sim\mathcal{N}(\mathbf{0},\sigma_w^2\mathbf{I}),
\qquad
\boldsymbol{\eta}_{i,t}^{a}\sim\mathcal{N}(\mathbf{0},\sigma_a^2\mathbf{I}),
\qquad
\boldsymbol{\eta}_{i,t}^{m}\sim\mathcal{N}(\mathbf{0},\sigma_m^2\mathbf{I}).
\label{eq:process_noise_distributions}
\end{equation}

Observation uncertainty is applied to position, speed, heading, flight-path angle, and energy. The observed self-position and energy are

\begin{equation}
\hat{\mathbf{p}}_{i,t}
=
\mathbf{p}_{i,t}
+
\boldsymbol{\epsilon}_{i,t}^{p},
\qquad
\hat{E}_{i,t}
=
E_{i,t}\left(1+\epsilon_{i,t}^{E}\right),
\label{eq:self_observations}
\end{equation}

with

\begin{equation}
\boldsymbol{\epsilon}_{i,t}^{p}
\sim
\mathcal{N}
\left(
\mathbf{0},
\sigma_{p,\mathrm{self}}^{2}\mathbf{I}
\right).
\label{eq:self_position_noise_distribution}
\end{equation}

The observed velocity is reconstructed from noisy speed and orientation measurements:

\begin{equation}
\hat{\mathbf{v}}_{i,t}
=
\hat{V}_{i,t}
\begin{bmatrix}
\cos(\hat{\gamma}_{i,t})\cos(\hat{\psi}_{i,t})\\
\cos(\hat{\gamma}_{i,t})\sin(\hat{\psi}_{i,t})\\
\sin(\hat{\gamma}_{i,t})
\end{bmatrix}.
\label{eq:observed_velocity_vector}
\end{equation}

For neighboring aircraft, communication delay is modeled as

\begin{equation}
\mathbf{p}_{j,t}^{\mathrm{delay}}
=
\mathbf{p}_{j,t}
-
\hat{\mathbf{v}}_{j,t}\tau_{ij,t},
\qquad
\tau_{ij,t}\sim\mathcal{U}(0,0.5),
\label{eq:delayed_neighbor_position}
\end{equation}

and the observed position is

\begin{equation}
\hat{\mathbf{p}}_{j,t}
=
\mathbf{p}_{j,t}^{\mathrm{delay}}
+
\boldsymbol{\epsilon}_{j,t}^{p}.
\label{eq:neighbor_position_noise}
\end{equation}

Neighbor information is removed with probability

\begin{equation}
P_{\mathrm{drop}}=0.05.
\label{eq:dropout_probability}
\end{equation}

These uncertainty terms provide noisy, delayed, and incomplete observations for MARL decision-making.

\vspace{0.2cm}
\noindent \textbf{Energy Consumption Model}
\label{subsec:energy_consumption_model}

The remaining energy of aircraft $i$ is updated as

\begin{equation}
E_{i,t+1}
=
E_{i,t}
-
P_{i,t}\Delta t,
\label{eq:energy_update}
\end{equation}

where $P_{i,t}$ is the power associated with the selected maneuver. Forward, vertical, and turning flight are modeled separately.

For forward flight,

\begin{equation}
P_{\mathrm{forward}}
=
P_1+P_2+P_3+P_4,
\label{eq:forward_power_total}
\end{equation}

where

\begin{equation}
P_1
=
\frac{W^{3/2}}{\sqrt{n\rho A}}
\frac{\delta}{8}sC_T^{-3/2},
\qquad
P_2
=
\frac{3}{8}\delta sV^2
\sqrt{\frac{W_r\rho A}{C_T}},
\label{eq:forward_power_terms12}
\end{equation}

\begin{equation}
P_3
=
(1+k)
\frac{W^{3/2}}{\sqrt{2n\rho A}}
\sqrt{
\sqrt{1+\frac{V^4}{4v_0^4}}
-
\frac{V^2}{2v_0^2}
},
\qquad
P_4
=
\frac{n}{2}S_{\mathrm{fp,par}}\rho V^3.
\label{eq:forward_power_terms34}
\end{equation}

For vertical flight, the hover power is

\begin{equation}
P_{\mathrm{hover}}
=
P_{\mathrm{blade}}
+
P_{\mathrm{induced}},
\label{eq:hover_power}
\end{equation}

with

\begin{equation}
P_{\mathrm{blade}}
=
\frac{W^{3/2}}{\sqrt{n\rho A}}
\frac{\delta}{8}sC_T^{-3/2},
\qquad
P_{\mathrm{induced}}
=
(1+k)\frac{W^{3/2}}{\sqrt{2n\rho A}}.
\label{eq:hover_power_terms}
\end{equation}

The climb and descent power are

\begin{equation}
\begin{aligned}
P_{\mathrm{vertical}}^{\mathrm{up}}
&=
P_{\mathrm{hover}}
+\frac{1}{2}WV
+\frac{n}{4}S_{\mathrm{fp,perp}}\rho V^3 \\
&\quad+
\left(
\frac{W}{2}
+\frac{n}{4}S_{\mathrm{fp,perp}}\rho V^2
\right)
\sqrt{
\left(
1+\frac{S_{\mathrm{fp,perp}}}{A}
\right)V^2
+\frac{2W}{n\rho A}
},
\end{aligned}
\label{eq:vertical_power_up}
\end{equation}

\begin{equation}
\begin{aligned}
P_{\mathrm{vertical}}^{\mathrm{down}}
&=
P_{\mathrm{hover}}
+\frac{1}{2}WV
-\frac{n}{4}S_{\mathrm{fp,perp}}\rho V^3 \\
&\quad+
\left(
\frac{W}{2}
-\frac{n}{4}S_{\mathrm{fp,perp}}\rho V^2
\right)
\sqrt{
\left(
1-\frac{S_{\mathrm{fp,perp}}}{A}
\right)V^2
+\frac{2W}{n\rho A}
}.
\end{aligned}
\label{eq:vertical_power_down}
\end{equation}

For turning flight, the total power is obtained by summing the power of all rotors:

\begin{equation}
P_{\mathrm{turn}}
=
\sum_{r=1}^{n}
\left(
P_{1,r}+P_{2,r}+P_{3,r}+P_{4,r}
\right),
\label{eq:turn_power_total}
\end{equation}

where

\begin{equation}
P_{1,r}
=
\frac{\delta}{8}
\rho^{-1/2}sA^{-1/2}C_T^{-3/2}W_r^{3/2},
\qquad
P_{2,r}
=
\frac{3}{8}\delta sV^2
\sqrt{\frac{W_r\rho A}{C_T}},
\label{eq:turn_power_terms12}
\end{equation}

\begin{equation}
P_{3,r}
=
(1+k)
\frac{W_r^{3/2}}{\sqrt{2\rho A}}
\sqrt{
\sqrt{1+\frac{V^4}{4v_0^4}}
-
\frac{V^2}{2v_0^2}
},
\qquad
P_{4,r}
=
\frac{1}{2}S_{\mathrm{fp,par}}\rho V^3.
\label{eq:turn_power_terms34}
\end{equation}

The power used in Equation~\eqref{eq:energy_update} is selected according to the maneuver:

\begin{equation}
P_{i,t}
=
\begin{cases}
P_{\mathrm{forward}}, & \text{maintain cruise or speed control},\\
P_{\mathrm{turn}}, & \text{turn left or right},\\
P_{\mathrm{vertical}}^{\mathrm{up}}, & \text{climb},\\
P_{\mathrm{vertical}}^{\mathrm{down}}, & \text{descend or land}.
\end{cases}
\label{eq:power_action_selection}
\end{equation}

This formulation captures the different energy requirements of forward, turning, vertical, and landing maneuvers.

\subsection{MARL Decision-Making Model}
\label{subsec:marl_decision_making_framework}

Each aircraft is modeled as an autonomous agent that observes the local environment, selects an action, receives a reward, and updates its policy. Separate DQN networks are used for UAV and eVTOL agents to represent their different operating characteristics.

\subsubsection{State Space}
\label{par:marl_state_space}

The state of aircraft $i$ includes its own position, velocity, and remaining energy:

\begin{equation}
\mathbf{s}_{i,t}^{\mathrm{self}}
=
\begin{bmatrix}
x_{i,t} &
y_{i,t} &
z_{i,t} &
v_{x,i,t} &
v_{y,i,t} &
v_{z,i,t} &
E_{i,t}
\end{bmatrix}^{T}.
\label{eq:self_state}
\end{equation}

For neighboring aircraft $j$, the observed state is

\begin{equation}
\mathbf{s}_{ij,t}^{\mathrm{nbr}}
=
\begin{bmatrix}
\Delta x_{ij,t} &
\Delta y_{ij,t} &
\Delta z_{ij,t} &
v_{x,j,t} &
v_{y,j,t} &
v_{z,j,t} &
\tau_{ij,t}
\end{bmatrix}^{T},
\label{eq:neighbor_state}
\end{equation}

where

\begin{equation}
\Delta x_{ij,t}=x_{j,t}-x_{i,t},
\qquad
\Delta y_{ij,t}=y_{j,t}-y_{i,t},
\qquad
\Delta z_{ij,t}=z_{j,t}-z_{i,t},
\label{eq:relative_position_components}
\end{equation}

and $\tau_{ij,t}$ is the age of the neighboring-aircraft information.

The local neighbor set is

\begin{equation}
\mathcal{N}_{i,t}
=
\left\{
j\neq i:
\left\|
\mathbf{p}_{j,t}-\mathbf{p}_{i,t}
\right\|
\leq R_{\mathrm{local}}
\right\},
\qquad
R_{\mathrm{local}}=1500~\mathrm{m}.
\label{eq:neighbor_set}
\end{equation}

Each agent observes the three nearest neighbors. Missing entries are zero-padded. The complete state vector is

\begin{equation}
\mathbf{S}_{i,t}
=
\begin{bmatrix}
\mathbf{s}_{i,t}^{\mathrm{self}} \\
\mathbf{s}_{i1,t}^{\mathrm{nbr}} \\
\mathbf{s}_{i2,t}^{\mathrm{nbr}} \\
\mathbf{s}_{i3,t}^{\mathrm{nbr}}
\end{bmatrix},
\label{eq:complete_state}
\end{equation}

with input dimension

\begin{equation}
d_s=7+3(7)=28.
\label{eq:state_dimension}
\end{equation}

\subsubsection{Action Space}
\label{par:action_space}

Each aircraft selects one discrete action at each simulation time step. The updated action space contains 14 actions:

\begin{equation}
\mathcal{A}=\{0,1,\ldots,13\}.
\label{eq:action_space}
\end{equation}

The action set includes maintaining cruise, turning, performing a vertical maneuver, initiating a landing action, and changing the current speed by a specified percentage. Table~\ref{tab:action_space} summarizes the action definitions.

\begin{table}[h]
\centering
\caption{Discrete action space for each aircraft agent.}
\label{tab:action_space}
\begin{tabular}{cl}
\hline
\textbf{Action} & \textbf{Description} \\
\hline
1  & Maintain current heading and speed \\
2  & Turn left \\
3 & Turn right \\
4  & Vertical maneuver \\
5  & Emergency landing \\
6  & Controlled landing \\
7  & Decrease current speed by 5\% \\
8  & Decrease current speed by 10\% \\
9  & Decrease current speed by 15\% \\
10  & Decrease current speed by 20\% \\
11 & Increase current speed by 5\% \\
12 & Increase current speed by 10\% \\
13 & Increase current speed by 15\% \\
14 & Increase current speed by 20\% \\
\hline
\end{tabular}
\end{table}

The vertical maneuver corresponds to climbing for eVTOL aircraft and descending for UAVs. Turning actions modify the aircraft heading, while vertical actions modify the flight-path angle. For a speed-control action, the updated aircraft speed is

\begin{equation}
V_{i,t}^{a}
=
V_{i,t}\left(1+\alpha_a\right),
\label{eq:speed_control_action}
\end{equation}

where

\begin{equation}
\alpha_a
\in
\{-0.05,-0.10,-0.15,-0.20,
0.05,0.10,0.15,0.20\}.
\label{eq:speed_control_rates}
\end{equation}

The corresponding post-action velocity is

\begin{equation}
\mathbf{v}_{i,t}^{a}
=
V_{i,t}^{a}
\frac{\mathbf{v}_{i,t}}
{\left\|\mathbf{v}_{i,t}\right\|},
\label{eq:speed_control_velocity}
\end{equation}

so that the aircraft speed changes while its current travel direction remains unchanged. Landing actions end normal flight operation.

\subsubsection{Safety Metrics and Minimum-Separation Definition}
\label{par:safety_metrics_minimum_separation}

Safety is evaluated using LOS and collision. For aircraft pair $(i,j)$ at time step $t$, the relative distance is

\begin{equation}
d_{ij,t}
=
\left\|
\mathbf{p}_{j,t}
-
\mathbf{p}_{i,t}
\right\|.
\label{eq:aircraft_pair_distance}
\end{equation}

The pairwise LOS indicator is

\begin{equation}
L_{ij,t}
=
\begin{cases}
1, & d_{ij,t}<D_{\zeta_i}^{\min},\\
0, & d_{ij,t}\geq D_{\zeta_i}^{\min},
\end{cases}
\label{eq:pairwise_los_indicator}
\end{equation}

where

\begin{equation}
D_{\zeta_i}^{\min}
=
\begin{cases}
D_{\mathrm{eVTOL}}^{\min}, & \zeta_i=\mathrm{eVTOL},\\
D_{\mathrm{UAV}}^{\min}, & \zeta_i=\mathrm{UAV}.
\end{cases}
\label{eq:type_specific_minimum_separation}
\end{equation}

The number of LOS violations involving aircraft $i$ is

\begin{equation}
L_{i,t}
=
\sum_{\substack{j=1\\j\neq i}}^{N}
L_{ij,t}.
\label{eq:agent_los_count}
\end{equation}

A collision occurs when the pairwise distance falls below the collision threshold $D^{\mathrm{col}}$:

\begin{equation}
C_{ij,t}
=
\begin{cases}
1, & d_{ij,t}<D^{\mathrm{col}},\\
0, & d_{ij,t}\geq D^{\mathrm{col}},
\end{cases}
\qquad
D^{\mathrm{col}}<D_{\zeta_i}^{\min}.
\label{eq:pairwise_collision_indicator}
\end{equation}

The number of collisions involving aircraft $i$ is

\begin{equation}
C_{i,t}
=
\sum_{\substack{j=1\\j\neq i}}^{N}
C_{ij,t}.
\label{eq:agent_collision_count}
\end{equation}

During training, the minimum-separation thresholds were sampled as

\begin{equation}
D_{\mathrm{eVTOL}}^{\min},
D_{\mathrm{UAV}}^{\min}
\sim
\mathcal{U}(150,350)\ \mathrm{m}.
\label{eq:training_minimum_separation}
\end{equation}

These metrics are used in the reward function to penalize unsafe aircraft interactions.

\subsubsection{Reward Function}
\label{par:reward_function}

The reward function encourages each aircraft to move toward its assigned destination, remain within its assigned lane, and avoid loss-of-separation and collision events. For aircraft $i$ at time step $t$, the total reward is

\begin{equation}
R_{i,t}
=
R_{i,t}^{\mathrm{lat}}
+
R_{i,t}^{\mathrm{lane}}
-
P_{i,t}^{\mathrm{LOS}}
-
P_{i,t}^{\mathrm{col}}.
\label{eq:reward_general}
\end{equation}

The destination-direction component measures whether the aircraft velocity is aligned with the direction toward its assigned destination. Let $\mathbf{d}_i$ denote the unit direction vector toward the destination, and let

\begin{equation}
\hat{\mathbf{v}}_{i,t}
=
\frac{\mathbf{v}_{i,t}}
{\left\|\mathbf{v}_{i,t}\right\|}
\label{eq:unit_velocity}
\end{equation}

denote the unit velocity vector. The directional alignment value is

\begin{equation}
\alpha_{i,t}
=
\hat{\mathbf{v}}_{i,t}\cdot\mathbf{d}_i.
\label{eq:destination_alignment}
\end{equation}

The corresponding reward term is

\begin{equation}
R_{i,t}^{\mathrm{lat}}
=
1.5\alpha_{i,t}.
\label{eq:lateral_reward}
\end{equation}

A positive value indicates movement toward the destination, while a negative value indicates movement away from it.

Lane containment is represented by

\begin{equation}
I_{i,t}^{\mathrm{lane}}
=
\begin{cases}
1, & \text{if aircraft } i \text{ remains inside its assigned lane},\\
0, & \text{otherwise}.
\end{cases}
\label{eq:lane_indicator}
\end{equation}

The lane-containment reward is

\begin{equation}
R_{i,t}^{\mathrm{lane}}
=
0.2I_{i,t}^{\mathrm{lane}}
-
1.0\left(1-I_{i,t}^{\mathrm{lane}}\right).
\label{eq:lane_reward}
\end{equation}

Let $L_{i,t}$ denote the number of loss-of-separation events involving aircraft $i$, and let $C_{i,t}$ denote the number of collision events. The corresponding penalties are

\begin{equation}
P_{i,t}^{\mathrm{LOS}}
=
10L_{i,t},
\qquad
P_{i,t}^{\mathrm{col}}
=
30C_{i,t}.
\label{eq:safety_penalties}
\end{equation}

The complete reward function is therefore

\begin{equation}
\begin{aligned}
R_{i,t}
=
&\ 1.5\alpha_{i,t}
+
0.2I_{i,t}^{\mathrm{lane}}
-
1.0\left(1-I_{i,t}^{\mathrm{lane}}\right) -
10L_{i,t}
-
30C_{i,t}.
\end{aligned}
\label{eq:expanded_reward}
\end{equation}

This formulation rewards movement toward the assigned destination and operation within the assigned lane while penalizing loss-of-separation and collision events.

\subsubsection{Deep Q-Network Architecture.}
\label{par:dqn_architecture}

The decision-making policy for each aircraft is represented using a Deep Q-Network. Two separate DQN policies are used: one for UAV agents and one for eVTOL agents. This separation allows each aircraft category to use a policy structure associated with its own motion characteristics, speed range, energy model, and operating layer.

The Q-value function approximated by the DQN is denoted as

\begin{equation}
Q(\mathbf{S}_{i,t},a_{i,t};\theta),
\label{eq:q_function}
\end{equation}

\noindent where $\mathbf{S}_{i,t}$ is the state of aircraft $i$, $a_{i,t}$ is the selected action, and $\theta$ represents the neural network parameters. Each DQN receives the $28$-dimensional state vector as input and outputs one Q-value for each of the six discrete actions. The network architecture is summarized in Table~\ref{tab:dqn_architecture}.

\begin{table}[h]
\centering
\caption{DQN architecture used for UAV and eVTOL policy networks.}
\label{tab:dqn_architecture}
\begin{tabular}{lc}
\hline
\textbf{Layer} & \textbf{Number of Neurons} \\
\hline
Input layer & $28$ \\
Hidden layer 1 & $128$ \\
Hidden layer 2 & $128$ \\
Output layer & $6$ \\
\hline
\end{tabular}
\end{table}

\noindent The hidden layers use rectified linear unit activation functions. The network mapping can be expressed as

\begin{equation}
\mathbf{q}_{i,t}
=
f_{\theta}
\left(
\mathbf{S}_{i,t}
\right),
\label{eq:dqn_mapping}
\end{equation}

\noindent where $\mathbf{q}_{i,t}$ is the vector of Q-values produced by the DQN for aircraft $i$ at time step $t$:

\begin{equation}
\mathbf{q}_{i,t}
=
\begin{bmatrix}
Q(\mathbf{S}_{i,t},0;\theta) &
Q(\mathbf{S}_{i,t},1;\theta) &
\cdots &
Q(\mathbf{S}_{i,t},5;\theta)
\end{bmatrix}^{T}.
\label{eq:q_value_vector}
\end{equation}

\noindent During policy execution, the greedy action is selected as the action with the maximum Q-value:

\begin{equation}
a_{i,t}^{*}
=
\arg\max_{a\in\mathcal{A}}
Q(\mathbf{S}_{i,t},a;\theta).
\label{eq:greedy_action}
\end{equation}

\noindent The training process used to update the policy-network parameters is described in the following section.

\subsubsection{Training and Testing Procedure}
\label{subsec:training_testing_procedure}

This section describes how the UAV and eVTOL DQN policies were trained and tested. The training procedure updates the policy-network parameters using experience replay and target-network updates. The testing procedure evaluates the trained policies on independently generated scenarios without additional learning.

\subsubsection{Training Procedure}
\label{par:training_procedure}

The MARL model was trained using a Deep Q-Network learning procedure with experience replay and target-network updates. Training was conducted separately for the UAV and eVTOL policy networks. During each episode, the environment was initialized with a randomized heterogeneous traffic scenario, and each aircraft agent selected actions based on its local state observation.

Let $N_{\mathrm{ep}}$ denote the total number of training episodes. In this study,

\begin{equation}
N_{\mathrm{ep}}
=
20{,}000.
\label{eq:num_training_episodes}
\end{equation}

\noindent Let $e$ denote the training episode index, where

\begin{equation}
e
=
1,2,\ldots,N_{\mathrm{ep}}.
\label{eq:training_episode_index}
\end{equation}

\noindent Each episode consisted of a fixed number of simulation time steps:

\begin{equation}
T
=
50.
\label{eq:num_timesteps}
\end{equation}

\noindent At each time step, every active aircraft observes its state, selects an action, updates its position and velocity, receives a reward, and stores the resulting transition in the replay buffer.

During training, an $\epsilon$-greedy policy was used to balance exploration and exploitation. The action selected by aircraft $i$ at time step $t$ is defined as

\begin{equation}
a_{i,t}
=
\begin{cases}
\mathrm{random\ action}, & \mathrm{with\ probability}\ \epsilon,\\
\arg\max_{a\in\mathcal{A}} Q(\mathbf{S}_{i,t},a;\theta), & \mathrm{with\ probability}\ 1-\epsilon.
\end{cases}
\label{eq:epsilon_greedy}
\end{equation}

\noindent The exploration rate decays over training episodes according to

\begin{equation}
\epsilon_e
=
\max
\left(
\epsilon_{\min},
\epsilon_0\beta^{e}
\right),
\label{eq:epsilon_decay}
\end{equation}

\noindent where $\epsilon_e$ is the exploration rate in episode $e$, $\epsilon_0$ is the initial exploration rate, $\epsilon_{\min}$ is the minimum exploration rate, and $\beta$ is the decay factor.

For aircraft $i$ at time step $t$, the transition stored in the replay buffer is represented as

\begin{equation}
\left(
\mathbf{S}_{i,t},
a_{i,t},
r_{i,t},
\mathbf{S}_{i,t+1},
d_{i,t}
\right).
\label{eq:replay_transition}
\end{equation}

\noindent where $\mathbf{S}_{i,t}$ is the current state, $a_{i,t}$ is the selected action, $r_{i,t}$ is the received reward, $\mathbf{S}_{i,t+1}$ is the next state, and $d_{i,t}$ is a terminal indicator.

The replay buffer stores past transitions and is used to sample mini-batches during training. The replay buffer capacity was set to

\begin{equation}
C_{\mathrm{buffer}}
=
50{,}000.
\label{eq:replay_buffer_capacity}
\end{equation}

\noindent A mini-batch of transitions is randomly sampled from the replay buffer during each learning update. The batch size used in training was

\begin{equation}
B
=
64.
\label{eq:batch_size}
\end{equation}

\noindent Random mini-batch sampling reduces correlation among consecutive transitions and allows the neural network to learn from a mixture of past traffic states.

The DQN was trained by minimizing the temporal-difference error between the predicted Q-value and the Bellman target. For a sampled transition, the target value is computed as

\begin{equation}
y_{i,t}
=
r_{i,t}
+
\gamma
\max_{a'}
Q
\left(
\mathbf{S}_{i,t+1},
a';
\theta^{-}
\right).
\label{eq:dqn_training_target}
\end{equation}

\noindent where $\gamma$ is the discount factor, $\theta^{-}$ denotes the target-network parameters, and $a'$ denotes a candidate action in the next state.

If the transition corresponds to a terminal state, the target value is computed as

\begin{equation}
y_{i,t}
=
r_{i,t}.
\label{eq:terminal_target}
\end{equation}

\noindent This prevents the network from assigning future value to states that terminate the episode.

The loss function is defined as the squared temporal-difference error:

\begin{equation}
\mathcal{L}(\theta)
=
\left[
y_{i,t}
-
Q
\left(
\mathbf{S}_{i,t},
a_{i,t};
\theta
\right)
\right]^2.
\label{eq:dqn_training_loss}
\end{equation}

\noindent The policy-network parameters $\theta$ are updated by minimizing Equation~\eqref{eq:dqn_training_loss} using the Adam optimizer.

The learning rate used for optimization was

\begin{equation}
\alpha_{\mathrm{lr}}
=
10^{-4}.
\label{eq:learning_rate}
\end{equation}

\noindent The discount factor was set to

\begin{equation}
\gamma
=
0.99.
\label{eq:discount_factor}
\end{equation}

\noindent A separate target network was used to stabilize Q-learning updates. The target network parameters were updated periodically using the policy-network parameters. The target update interval was

\begin{equation}
N_{\mathrm{target}}
=
50~\mathrm{episodes}.
\label{eq:target_update_interval}
\end{equation}

\noindent Therefore, after every $50$ training episodes, the target network was synchronized with the current policy network.

The key training hyperparameters are summarized in Table~\ref{tab:training_hyperparameters}.

\begin{table}[h]
\centering
\caption{Training hyperparameters used for the DQN-based MARL model.}
\label{tab:training_hyperparameters}
\begin{tabular}{lc}
\hline
\textbf{Parameter} & \textbf{Value} \\
\hline
Training episodes & $20{,}000$ \\
Time steps per episode & $50$ \\
Batch size & $64$ \\
Learning rate & $10^{-4}$ \\
Discount factor & $0.99$ \\
Replay buffer capacity & $50{,}000$ \\
Target-network update interval & Every $50$ episodes \\
\hline
\end{tabular}
\end{table}

\noindent The training process allows the UAV and eVTOL agents to update their policies through repeated interaction with randomized traffic scenarios. Over the training episodes, the agents learn action-value estimates for maneuver selection under heterogeneous traffic, uncertainty, and corridor constraints.

\clearpage 

\subsubsection{Testing Procedure}
\label{par:testing_procedure}

The testing scenarios were generated using a grid of aircraft-density levels and minimum-separation thresholds. The aircraft-density thresholds were defined as

\begin{equation}
\mathcal{D}
=
\{4,6,8,10,12,14,16,18,20,24\}.
\label{eq:test_density_levels}
\end{equation}

\noindent The minimum-separation thresholds were defined as

\begin{equation}
\mathcal{S}
=
\{150,175,200,225,250,275,300,325,350\}.
\label{eq:test_separation_levels}
\end{equation}

\noindent Each testing scenario was created by combining one aircraft-density level from $\mathcal{D}$ with one base minimum-separation threshold from $\mathcal{S}$. The resulting scenario grid is summarized in Table~\ref{tab:test_scenario_grid}.

\begin{table}[h]
\centering
\caption{Testing scenario grid based on aircraft density levels and minimum-separation thresholds.}
\label{tab:test_scenario_grid}
\resizebox{\textwidth}{!}{
\begin{tabular}{c|ccccccccc}
\hline
\textbf{Number of Agents} &
\multicolumn{9}{c}{\textbf{Minimum Separation Threshold (m)}} \\
\cline{2-10}
&
\textbf{150} &
\textbf{175} &
\textbf{200} &
\textbf{225} &
\textbf{250} &
\textbf{275} &
\textbf{300} &
\textbf{325} &
\textbf{350} \\
\hline
4  & $(4,150)$  & $(4,175)$  & $(4,200)$  & $(4,225)$  & $(4,250)$  & $(4,275)$  & $(4,300)$  & $(4,325)$  & $(4,350)$  \\
6  & $(6,150)$  & $(6,175)$  & $(6,200)$  & $(6,225)$  & $(6,250)$  & $(6,275)$  & $(6,300)$  & $(6,325)$  & $(6,350)$  \\
8  & $(8,150)$  & $(8,175)$  & $(8,200)$  & $(8,225)$  & $(8,250)$  & $(8,275)$  & $(8,300)$  & $(8,325)$  & $(8,350)$  \\
10 & $(10,150)$ & $(10,175)$ & $(10,200)$ & $(10,225)$ & $(10,250)$ & $(10,275)$ & $(10,300)$ & $(10,325)$ & $(10,350)$ \\
12 & $(12,150)$ & $(12,175)$ & $(12,200)$ & $(12,225)$ & $(12,250)$ & $(12,275)$ & $(12,300)$ & $(12,325)$ & $(12,350)$ \\
14 & $(14,150)$ & $(14,175)$ & $(14,200)$ & $(14,225)$ & $(14,250)$ & $(14,275)$ & $(14,300)$ & $(14,325)$ & $(14,350)$ \\
16 & $(16,150)$ & $(16,175)$ & $(16,200)$ & $(16,225)$ & $(16,250)$ & $(16,275)$ & $(16,300)$ & $(16,325)$ & $(16,350)$ \\
18 & $(18,150)$ & $(18,175)$ & $(18,200)$ & $(18,225)$ & $(18,250)$ & $(18,275)$ & $(18,300)$ & $(18,325)$ & $(18,350)$ \\
20 & $(20,150)$ & $(20,175)$ & $(20,200)$ & $(20,225)$ & $(20,250)$ & $(20,275)$ & $(20,300)$ & $(20,325)$ & $(20,350)$ \\
24 & $(24,150)$ & $(24,175)$ & $(24,200)$ & $(24,225)$ & $(24,250)$ & $(24,275)$ & $(24,300)$ & $(24,325)$ & $(24,350)$ \\
\hline
\end{tabular}
}
\end{table}

\noindent Therefore, the total number of base testing scenarios was $N_{\mathrm{test}}
=
|\mathcal{D}|
\times
|\mathcal{S}|
=
10 \times 9
=
90.$ Thus, the trained policies were evaluated over $90$ base testing scenarios covering different combinations of traffic density and minimum-separation requirement. For each density level, the aircraft population was divided between eVTOL aircraft and UAVs. The number of eVTOL aircraft was assigned as one half of the total aircraft count using integer division, and the remaining aircraft were assigned as UAVs. This produced heterogeneous testing scenarios across all density levels.

Each base testing scenario was evaluated over $50$ independent test episodes, and each episode was simulated for $50$ time steps. For a given base scenario, the aircraft-density level and minimum-separation threshold were kept fixed across all $50$ repeated test episodes. The repeated episodes differed only in the randomly generated initial aircraft positions and spacing within the same corridor configuration. This allowed the trained policies to be evaluated under multiple initial traffic arrangements while preserving the same density and separation requirement for each base scenario.

Let $m$ denote the repeated test episode index for a given base scenario, where $m = 1,2,\ldots,50.$

\noindent For each base scenario defined by density level $d \in \mathcal{D}$ and minimum-separation threshold $s \in \mathcal{S}$, the total aircraft count and separation threshold remained fixed over the $50$ repeated test episodes. The aircraft initial positions were randomly generated for each repeated episode as

\begin{equation}
\mathbf{p}_{i,0}^{(m)}
=
\left(
x_{i,0}^{(m)},
y_{i,0}^{(m)},
z_{i,0}^{(m)}
\right),
\qquad
m=1,2,\ldots,50,
\label{eq:test_initial_position}
\end{equation}

\noindent where $\mathbf{p}_{i,0}^{(m)}$ represents the initial position of aircraft $i$ in repeated test episode $m$. The random initialization changes the starting locations and spacing among aircraft across repeated episodes, while the scenario-level density $d$ and minimum-separation threshold $s$ remain unchanged. Therefore, the repeated testing episodes evaluate the robustness of the trained policies to different initial traffic arrangements under the same base scenario configuration.

At the beginning of each testing scenario, the traffic environment was initialized using the same procedure used during training. However, the testing scenarios were generated independently from the training episodes using the structured scenario grid, so the trained policies were evaluated on unseen testing episodes rather than repeated training episodes. During testing, the trained UAV and eVTOL policy networks were used only for action selection. The eVTOL policy was used for eVTOL agents, and the UAV policy was used for UAV agents.

At each testing time step, each active aircraft observed its local state and selected the action with the highest Q-value among the allowed actions. The same aircraft dynamics, uncertainty model, energy-consumption model, corridor-containment check, safety metrics, and reward function defined in the previous sections were used to compute testing outcomes. Unlike training, no exploration-based learning, replay-buffer sampling, gradient update, backpropagation, or target-network update was performed during testing.

The testing procedure recorded the safety outcomes, aircraft maneuvering behavior, and time-step-level aircraft responses for each evaluated scenario. The recorded outputs included the total number of loss-of-separation events and collision events; the number of turning actions, vertical maneuvers, emergency landings, controlled landings, out-of-track events, and out-of-corridor events for UAV and eVTOL aircraft; and the aircraft position, velocity, remaining energy, selected action, observed safety risk, alignment value, corridor status, valid-exit status, and immediate reward at each time step. The procedure also recorded the scenario settings and aggregate performance outcomes, including aircraft count, separation threshold, accumulated reward, loss-of-separation count, collision count, maneuver counts, landing counts, out-of-track count, and out-of-corridor count. Therefore, the testing procedure evaluates the trained MARL policies without additional learning. The recorded outputs are used to assess how the trained UAV and eVTOL agents respond to heterogeneous traffic, separation constraints, uncertainty, and corridor-boundary conditions across the full testing scenario grid.

\section{Assumptions}
\label{subsec:marl_assumptions}

The MARL model was developed under several assumptions about the operating environment and aircraft behavior. The airspace was assumed to consist of a structured three-dimensional corridor with fixed altitude layers, lane boundaries, and traffic-flow directions. UAVs and eVTOL aircraft were assigned to separate operating layers and remained within designated lanes. Each aircraft was treated as an autonomous agent with access to its own state and information from a limited number of nearby aircraft rather than complete knowledge of the entire airspace. Aircraft motion was represented using simplified kinematic relationships and predefined maneuvering limits instead of full six-degree-of-freedom dynamics. The initial number of aircraft, positions, speeds, and energy levels were generated within the specified ranges, and aircraft energy decreased according to the selected maneuver.

Surveillance degradation was assumed to occur through observation noise, communication delay, information dropout, wind disturbance, actuator uncertainty, and model uncertainty. LOS and collision were defined using fixed distance thresholds, with the collision threshold smaller than the minimum-separation threshold. Each agent selected one action from the discrete 14-action space, including maintain, turning, vertical, landing, and speed-control actions. The reward function was assumed to represent the main operational objectives by rewarding destination-directed movement and lane containment and penalizing LOS and collision events. Training and testing were conducted over finite simulation horizons, and policy performance was evaluated across the defined aircraft-density and minimum-separation scenarios. Therefore, the findings apply to the operating conditions, aircraft characteristics, uncertainty levels, corridor geometry, and decision rules represented in the simulation.

\section{Solution Algorithm}

All experiments were conducted using Python on a workstation equipped with an NVIDIA RTX 6000 Ada Generation GPU. The MARL simulation environment, aircraft dynamics, uncertainty models, reward function, DQN-based learning framework, and result visualization procedures were implemented using Python libraries, including NumPy, Pandas, PyProj, PyTorch, Matplotlib, and tqdm. PyTorch was used to construct and train the Deep Q-Networks, and CUDA acceleration was used when available to support GPU-based training.

To improve reproducibility, a fixed random seed of 42 was used for Python's random module, NumPy, and PyTorch. When CUDA was available, the CUDA random seeds were also fixed. In addition, deterministic CuDNN settings were applied by setting \texttt{torch.backends.cudnn.deterministic = True} and \texttt{torch.backends.cudnn.benchmark = False}. These settings were used to reduce randomness in the training process and improve consistency across repeated experimental runs.

The detailed steps used to train and test the proposed MARL model are described in the methodology section. In summary, the aircraft agents learn conflict-resolution policies through repeated interaction with the simulation environment using a DQN-based learning procedure. After training, the learned policies were frozen, and no additional exploration or policy updates were performed during testing. The trained agents were then evaluated under unseen traffic-density and minimum-separation scenarios to assess their ability to maintain safe separation, reduce loss-of-separation events, avoid collisions, and preserve organized traffic flow under degraded surveillance conditions.

\section{Results}
\label{sec:marl_results}

This section presents the testing results of the trained MARL policies under degraded surveillance conditions. The policies were first evaluated across the full testing grid. Pareto analysis was then used to identify configurations that balance corridor capacity and safety. The selected Pareto-optimal cases were examined using time-step-level action analysis and LOS-duration analysis. Training convergence and repeated testing performance were also evaluated.

\subsection{Results for All Testing Scenarios}
\label{subsec:results_all_90_scenarios}

The trained MARL policies were evaluated across 90 testing scenarios generated from different aircraft-density levels and minimum-separation thresholds. Safety performance changed with both the number of agents and the required separation distance. Higher aircraft density produced more interactions among aircraft, while larger separation thresholds caused more aircraft pairs to be classified as loss-of-separation events.

Collisions were nearly eliminated across the testing grid. No collision occurred in any scenario except the case with 14 agents and a 350~m minimum-separation threshold, where one collision was recorded. No emergency landing or controlled landing action was selected. The agents relied on maintain, turning, vertical, and speed-control actions.

\begin{figure}[th]
\centering
\includegraphics[width=\textwidth]{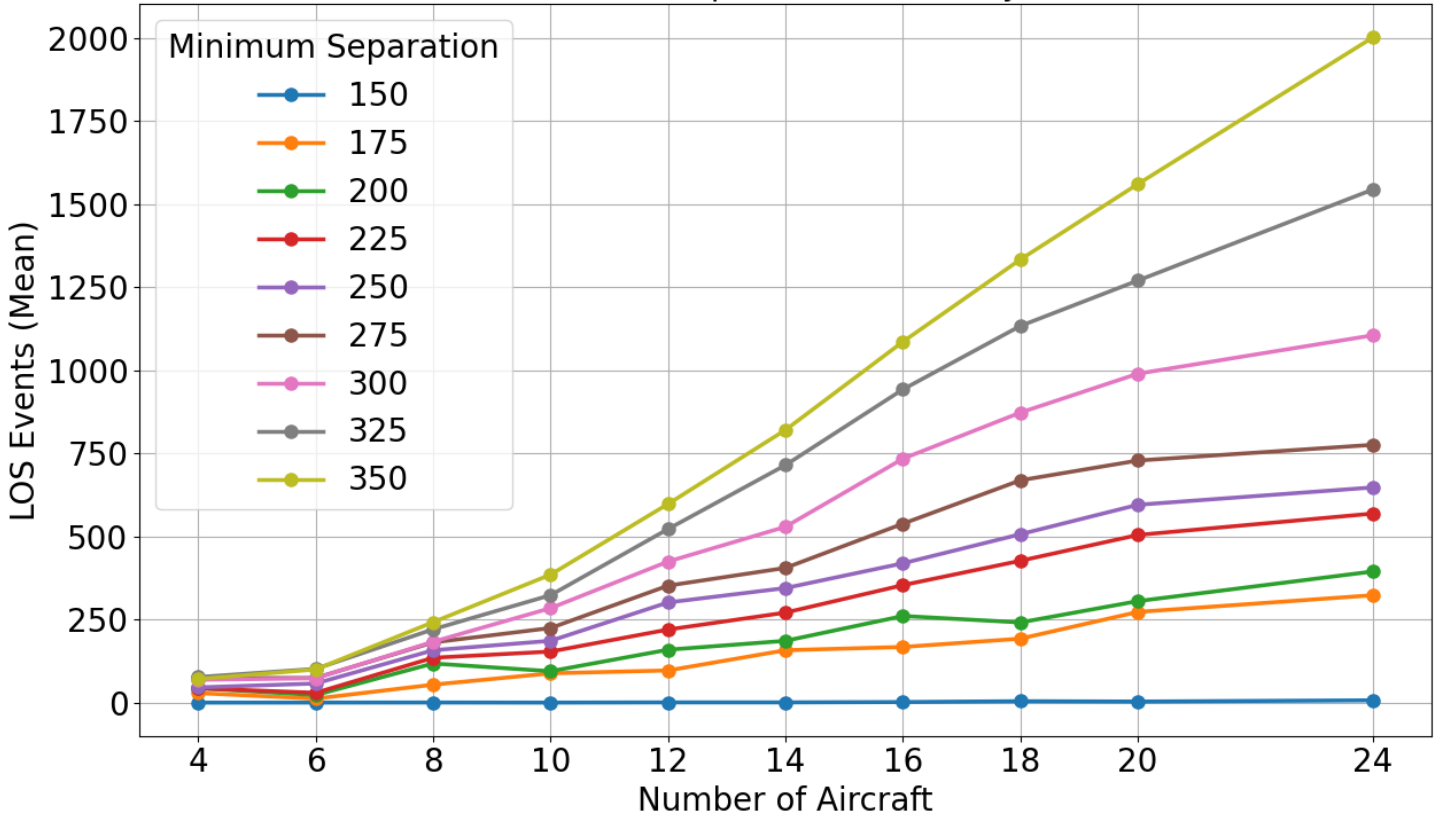}
\caption{Loss-of-separation events across all testing scenarios with different numbers of agents and minimum-separation thresholds.}
\label{figLOS}
\end{figure}

Figure~\ref{figLOS} presents the loss-of-separation results across the testing grid. The number of LOS events generally increased with aircraft density because more agents created more possible interactions within the corridor. LOS events also increased as the minimum-separation threshold became larger because a wider safety buffer caused more aircraft pairs to violate the required separation distance.

Although LOS events increased under denser traffic and larger separation thresholds, collisions remained nearly zero. Most unsafe encounters remained separation-buffer violations and did not progress to physical collisions.

The recorded action data showed that the agents used more corrective maneuvers as aircraft density and separation requirements increased. Turning, vertical, and speed-control actions were used to change relative motion and improve separation. Maintain actions were selected when no immediate correction was required. The absence of landing actions shows that the evaluated conflicts were handled through in-flight maneuvers.

Corridor-boundary violations were more closely related to aircraft density than to the minimum-separation threshold. As the number of agents increased, the aircraft faced more constrained interactions and performed more avoidance maneuvers, which increased the chance of leaving the assigned corridor.

The 90 testing scenarios revealed a trade-off between corridor capacity and safety. A larger number of agents increased corridor capacity but also increased LOS events and corrective maneuvers. Pareto analysis was therefore used to identify configurations that balanced aircraft count and safety performance.


\subsection{Pareto Analysis of Traffic Configurations}
\label{subsec:pareto_analysis}

After evaluating all $90$ testing scenarios, Pareto analysis was used to identify traffic configurations that provide the best trade-off between corridor capacity and safety. The full scenario grid shows that increasing the number of agents increases corridor capacity because more aircraft are handled within the same corridor. However, increasing the number of agents also increases traffic interaction complexity and tends to increase the number of loss-of-separation events. As a result, higher-capacity scenarios may reduce safety performance. This creates a trade-off between two competing objectives: maximizing capacity and maintaining safety.

In this study, the capacity score represents the ability of the corridor to accommodate more agents, while the safety score represents the ability of the trained policy to maintain safe separation. Since the capacity score increases with the number of agents, higher-density configurations generally receive higher capacity scores. In contrast, the safety score decreases when loss-of-separation events increase. Therefore, a configuration with many agents may have high capacity but lower safety, while a configuration with fewer agents may have high safety but lower capacity.

Pareto analysis was used to identify the scenarios that provide the best balance between these two objectives. A scenario is considered Pareto-optimal if no other tested scenario improves one objective without reducing the other objective. In other words, a Pareto-optimal configuration represents a non-dominated trade-off between safety and capacity.

\begin{figure}[tb!]
\centering
\includegraphics[width=\textwidth]{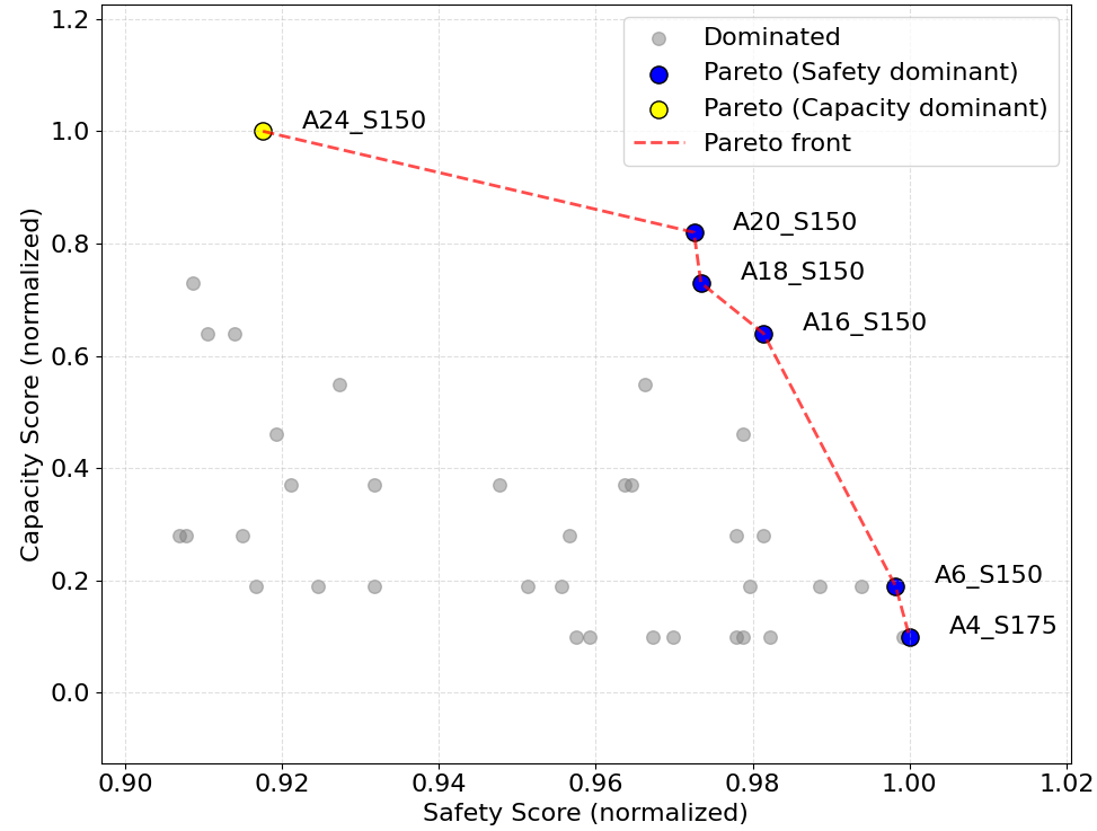}
\caption{Pareto analysis of the $90$ testing scenarios based on safety score and capacity score.}
\label{fig_pareto_optimal}
\end{figure}

Figure~\ref{fig_pareto_optimal} presents the Pareto analysis for the $90$ testing scenarios. The $x$-axis represents the safety score, which increases from left to right, and the $y$-axis represents the capacity score, which increases from bottom to top. The gray points represent dominated configurations. These are scenarios for which at least one other configuration provides better performance in both safety and capacity. The blue points represent safety-dominant configurations, which prioritize safety over capacity. The yellow points represent capacity-dominant configurations, which prioritize capacity over safety. The red line represents the Pareto front, which connects the Pareto-optimal scenarios.

The Pareto front identifies the configurations that provide the best achievable trade-offs between safety and capacity among the tested scenarios. The Pareto-optimal configurations include $4$ agents with a $175$ m minimum-separation threshold and higher-density configurations with $16$, $18$, $20$, and $24$ agents at a $150$ m minimum-separation threshold. These configurations are important because no other tested scenario can improve either safety or capacity without reducing the other objective.

Based on the Pareto analysis, the remaining evaluation focuses on the Pareto-optimal configurations rather than all $90$ testing scenarios. The next subsection compares agent action-selection behavior and examines how LOS frequency and duration vary across these configurations. A representative two-agent encounter is then presented to illustrate one example of LOS resolution.

\subsection{Action Selection and LOS Characteristics}
\label{subsec:timestep_action_los_pareto}

\begin{figure}[h]
\centering
\includegraphics[width=\textwidth]{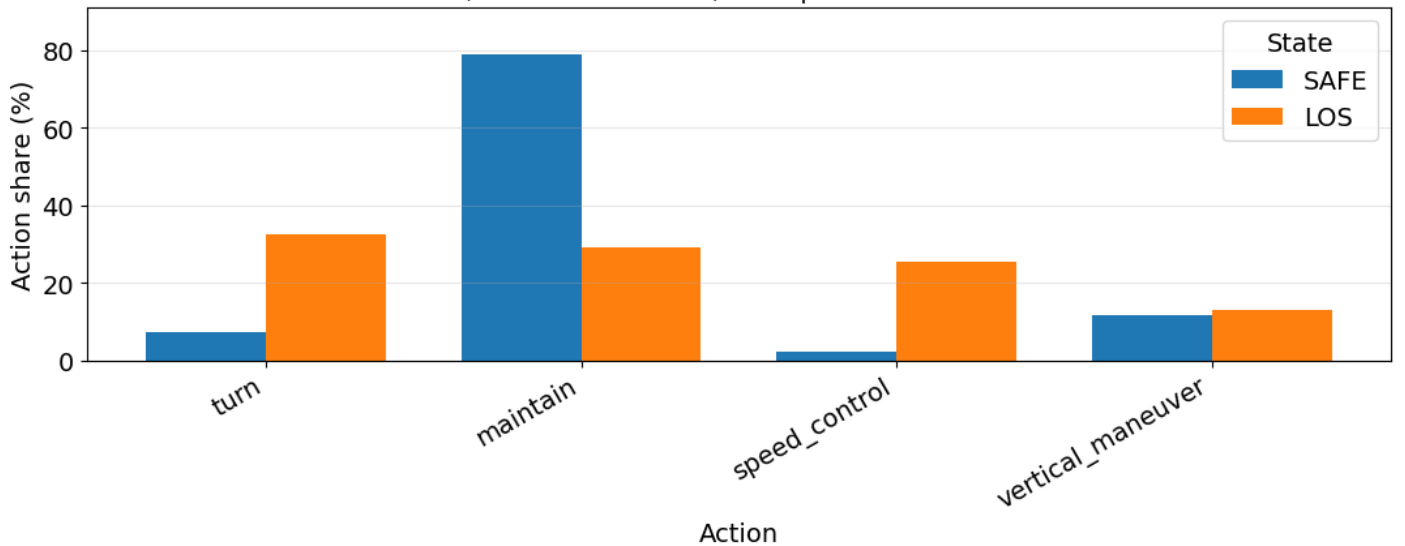}
\caption{Action shares during SAFE and LOS states across the evaluated test cases.}
\label{fig_action_selection}
\end{figure}

The action-selection analysis compares how the trained agents behaved during SAFE and LOS states across the evaluated test cases. A SAFE state represents a time step in which the required separation was maintained, whereas an LOS state represents a time step in which the distance between at least two aircraft fell below the applicable minimum-separation threshold. Figure~\ref{fig_action_selection} shows a change in agent behavior after an LOS event occurred. During SAFE states, the maintain action accounted for approximately 79\% of all selected actions. Turning accounted for about 7\%, vertical maneuvers for approximately 12\%, and speed-control actions for only about 2\%. This distribution shows that the agents generally preserved their current heading and speed when sufficient separation was available, while corrective actions were used infrequently. However, during LOS states, the agents shifted toward active conflict resolution. Turning became the most frequently selected action, accounting for approximately 33\% of the decisions. The maintain action decreased to approximately 29\%, while speed-control actions increased to about 25\%. Vertical maneuvers accounted for the remaining 13\%. The higher use of turning and speed-control actions shows that the agents primarily resolved conflicts by modifying horizontal motion. Vertical maneuvers were used less than turning and speed-control actions. The continued use of maintain actions during some LOS time steps does not indicate that no response occurred. An earlier turn or speed adjustment has already changed the encounter geometry, allowing the agents to maintain their new motion while separation gradually recovers.

\begin{figure}[p]
\centering
\includegraphics[width=\textwidth]{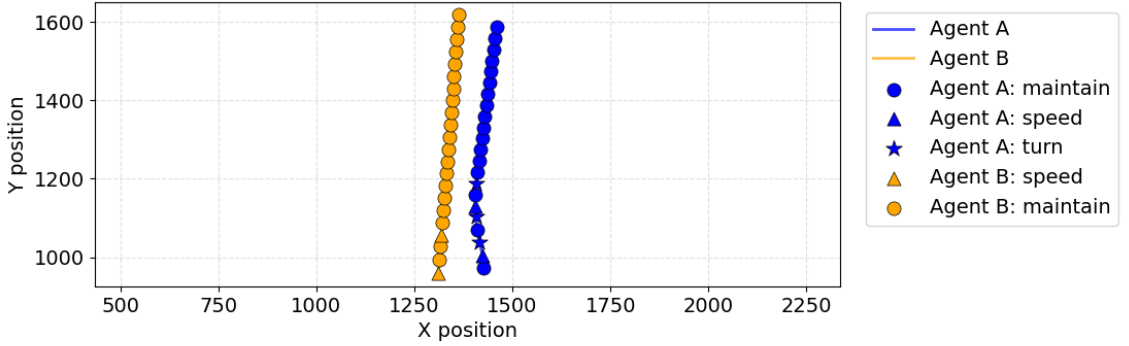}
\caption{Two-agent trajectories moving upward; circles indicate maintain, triangles speed control, and stars turning.}
\label{fig_los_trajectory}
\end{figure}

A representative two-agent encounter was examined to show how turn, speed-control, and maintain actions affected aircraft motion, speed, and pairwise separation. Figure~\ref{fig_los_trajectory} shows both aircraft moving upward, from lower to higher $y$ positions, along nearly parallel and closely spaced paths. Agent A traveled on the right side of the encounter, while Agent B traveled on the left. Without corrective action, the aircraft would have continued converging and could have reached the collision threshold within the next few time steps. Instead, both agents applied speed-control actions during the early stage of the encounter, and Agent A also selected turn actions that slightly changed its direction. After these corrective actions altered the relative motion, both aircraft mainly returned to the maintain action and continued upward along increasingly separated paths.

\begin{figure}[p]
\centering
\includegraphics[width=\textwidth]{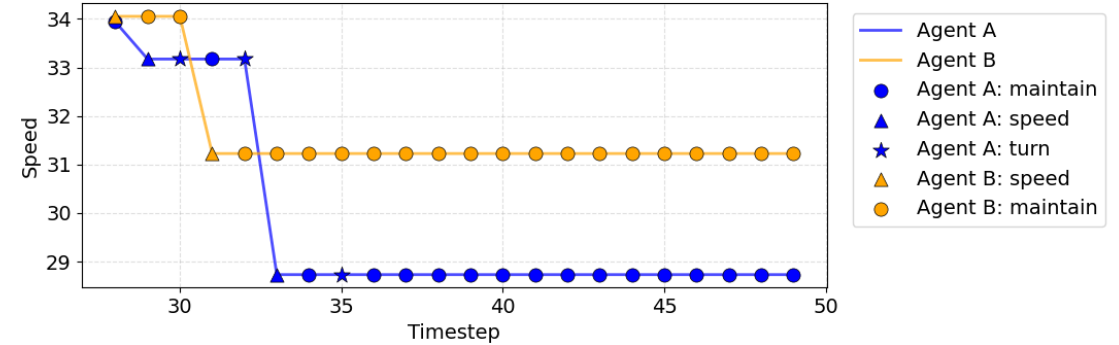}
\caption{Aircraft speeds and selected actions during the two-agent LOS-resolution case.}
\label{fig_los_speed}
\end{figure}

\begin{figure}[p]
\centering
\includegraphics[width=\textwidth]{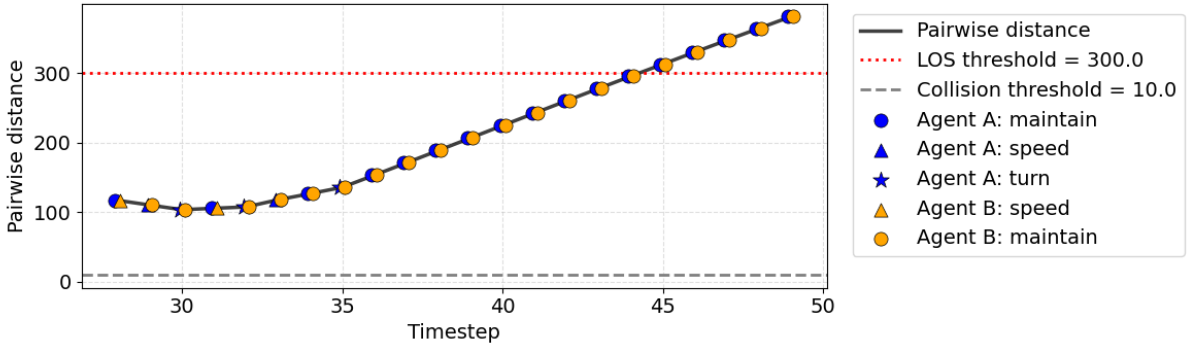}
\caption{Pairwise distance during LOS resolution, with minimum-separation and collision thresholds.}
\label{fig_los_distance}
\end{figure}

Figure~\ref{fig_los_speed} presents the corresponding speed profiles. Both aircraft initially traveled at approximately $34~\mathrm{m/s}$. Agent A first reduced its speed to approximately $33.2~\mathrm{m/s}$ and then selected turn actions while maintaining that speed. It later applied another speed reduction, decreasing its speed to approximately $28.7~\mathrm{m/s}$. Agent B maintained its initial speed during the first part of the encounter and then reduced its speed to approximately $31.2~\mathrm{m/s}$. After these adjustments, both agents maintained their new speeds for the remainder of the encounter. These coordinated actions changed the relative motion between the aircraft. Agent A applied the larger speed reduction and also changed heading, while Agent B made a smaller speed adjustment.

Figure~\ref{fig_los_distance} shows the resulting change in pairwise separation. At the beginning of the plotted encounter, the distance between the aircraft was approximately $118~\mathrm{m}$, which was below the $300~\mathrm{m}$ LOS threshold but well above the $10~\mathrm{m}$ collision threshold. The distance initially decreased to approximately $103~\mathrm{m}$, showing that the aircraft continued to converge during the early time steps. After the speed-control and turn actions were applied, the distance stopped decreasing and began to increase steadily. The pairwise separation exceeded the $300~\mathrm{m}$ threshold at approximately time step 45, marking the recovery of safe separation. No collision occurred because the distance remained well above the collision threshold throughout the encounter. The results show that LOS resolution resulted from the combined effect of heading and speed changes rather than from a single isolated maneuver. Agent A used turning and a larger speed reduction, while Agent B used a smaller speed reduction. Once these actions changed the encounter geometry, both agents returned to maintaining actions as separation continued to increase. This example shows how speed control complements turning by reducing the closing rate and restoring safe separation without requiring a vertical maneuver or landing action.

\begin{figure}[h]
\centering
\includegraphics[width=\textwidth]{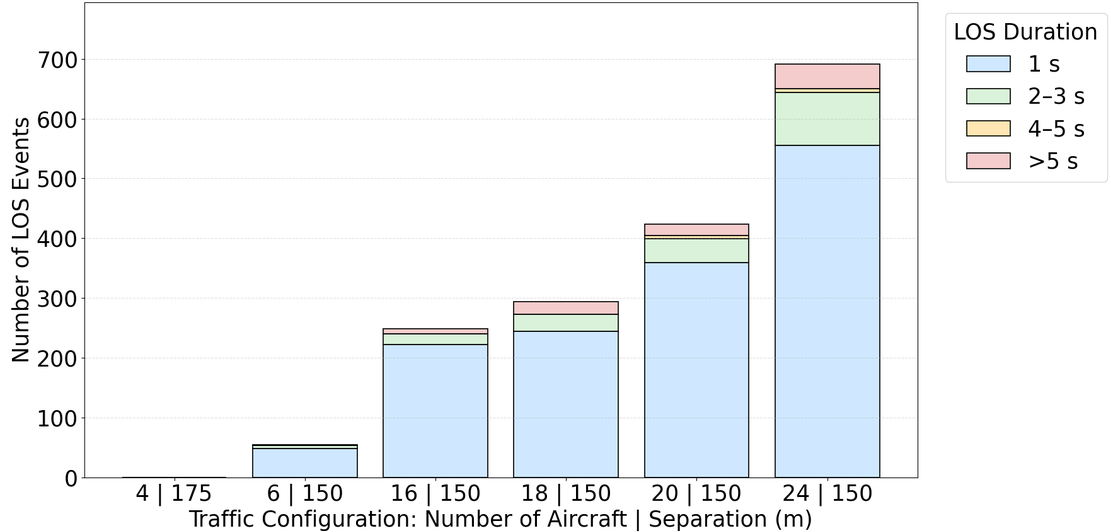}
\caption{LOS-event frequency and duration across selected traffic configurations, grouped into 1s, 2--3s, 4--5s, and more-than-5s categories.}
\label{fig_los_duration_frequency}
\end{figure}

Figure~\ref{fig_los_duration_frequency} compares both the frequency and duration of LOS events across the selected traffic configurations. The total height of each bar represents the total number of LOS events, while each colored section represents the number of events that lasted 1~s, 2--3~s, 4--5~s, or more than 5~s. As the number of aircraft increased, the bars became taller, showing that LOS events occurred more frequently in denser traffic. The 1-s category remained the largest in every case, which means that most LOS events were resolved quickly. However, the 2--3~s, 4--5~s, and more-than-5~s sections also became larger in the higher-density cases. This means that crowded traffic created not only more short LOS events, but also more extended events that required several time steps to resolve.

\subsection{Model Validation}
\label{subsec:model_validation_testing}


The MARL model was validated using training-convergence behavior and repeated testing on selected Pareto-optimal traffic configurations. The purpose of model validation was to examine whether the agents learned stable policies during training and whether the fixed trained policies produced consistent performance during testing. Since this is a simulation-based reinforcement learning model, validation does not imply certification-level proof of real-world safety. Rather, it provides evidence that the trained policies behave consistently under the modeled assumptions, reward structure, and testing conditions.

\subsubsection{Training Convergence Check}

\begin{figure}[tb!h]
\centering
\includegraphics[width=13cm,height=7.5cm]{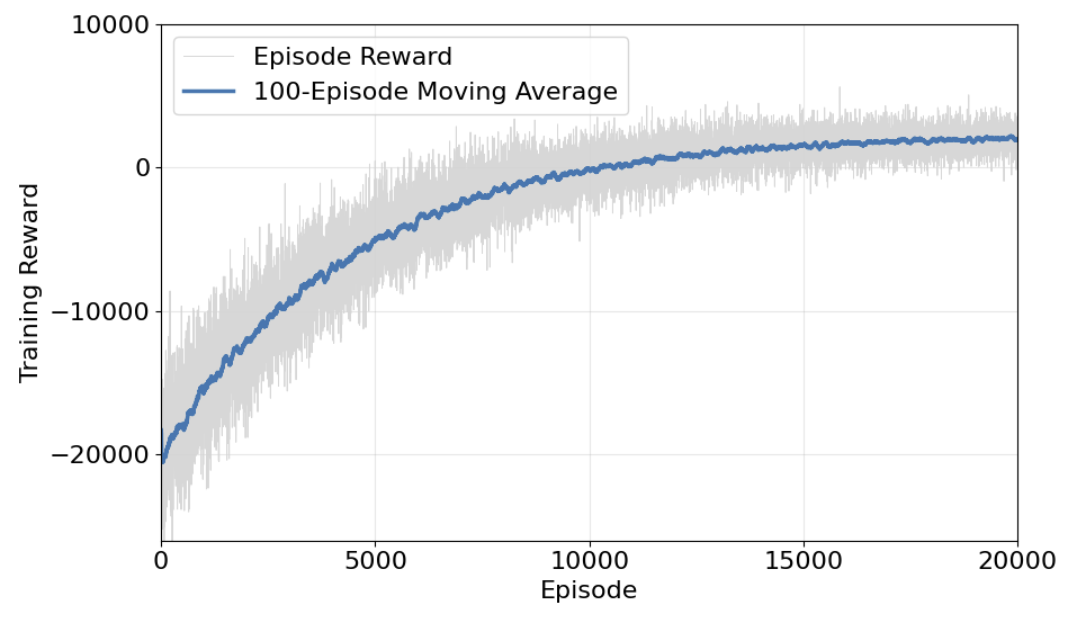}
\caption{Training reward over $20{,}000$ episodes with a $100$-episode moving average.}
\label{training_plot}
\end{figure}

Training performance was evaluated over $20{,}000$ episodes using both the raw episode reward and a $100$-episode moving average.  The training trend is shown in Figure~\ref{training_plot}. The raw reward fluctuated because the agents encountered different traffic conditions, aircraft configurations, and uncertainty realizations during training. These short-term variations made it difficult to assess the overall learning trend from the raw values alone. Therefore, the moving-average reward was used to provide a clearer indication of whether the agents were continuing to improve or had reached a stable level.

At the beginning of training, the moving-average reward was approximately $-20{,}000$, indicating that the agents frequently selected actions that resulted in unsafe or inefficient outcomes. As training progressed, the moving-average reward increased steadily and eventually reached a stable positive range. This upward trend shows that the agents gradually learned action-selection policies that better satisfied the safety and operational objectives represented in the reward function.

Convergence was evaluated near the end of training by comparing the $100$-episode moving-average reward at episodes $19{,}990$ and $20{,}000$. The relative change between these two values was approximately $2.76\%$. Because this change was below the selected $3\%$ convergence threshold, the training process was considered converged. This means that the smoothed reward changed only slightly during the final stage of training and that additional episodes were no longer producing substantial improvement in the learned policies. The convergence result confirms that the training reward had reached a stable range, but it does not by itself demonstrate that the policies were safe under all operating conditions. Since the reward combines several safety and operational terms, the stabilized reward provides evidence of stable learning. The safety effectiveness of the trained policies was therefore evaluated separately through repeated testing, LOS and collision metrics, and action-selection analysis across different traffic configurations.

\newpage
\subsubsection{Testing Stability Check}

After training, the UAV and eVTOL policy networks were fixed and evaluated without additional learning. During testing, no exploration-based learning, replay-buffer sampling, gradient update, backpropagation, or target-network update was performed. Therefore, the testing results represent the performance of the trained policies rather than continued training. The overall testing process first evaluated the trained policies across the full $90$-scenario grid. From these scenarios, six Pareto-optimal traffic configurations were identified as the best trade-off solutions between safety and capacity. 

For each selected Pareto-optimal traffic configuration, the trained policies were evaluated over $50$ independent test episodes, and each episode was simulated for $50$ time steps. For a given configuration, the total number of agents and the minimum-separation threshold used to define loss of separation remained fixed across the $50$ repeated episodes. However, the initial aircraft positions and spacing were randomly generated in each episode. This repeated-testing design was used to examine whether the fixed trained policies produced stable reward performance under different traffic arrangements for the same agent-count and separation-threshold setting.

\begin{figure}[h]
\centering
\includegraphics[width=\textwidth]{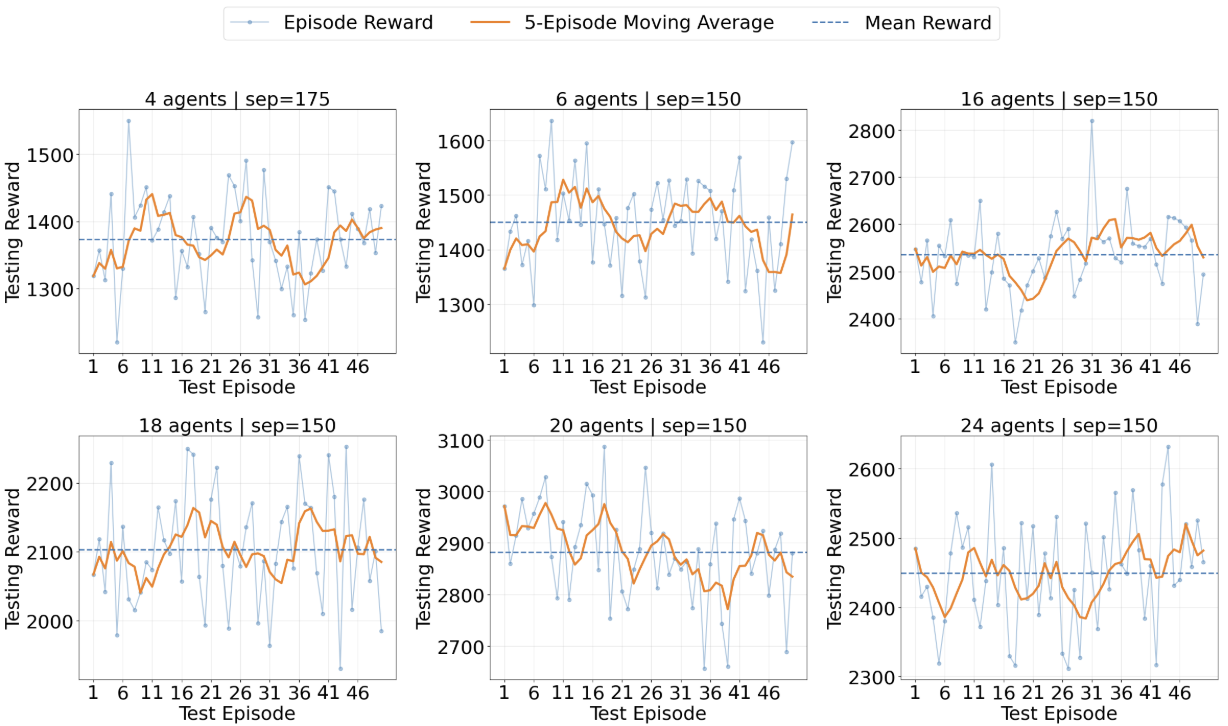}
\caption{Testing reward over $50$ independent test episodes for the six Pareto-optimal traffic configurations.}
\label{fig_test_reward}
\end{figure}

Unlike the training reward, the testing reward is not expected to show an increasing convergence trend because the policy networks are fixed during testing. Instead, testing stability was evaluated by examining whether the reward remained relatively consistent across repeated randomized episodes. Figure~\ref{fig_test_reward} shows that the testing reward curves remain approximately stable over the $50$ testing episodes for each selected Pareto-optimal configuration. In the figure, the episode reward represents the reward obtained in each individual test episode, while the moving-average curve represents the $5$-episode moving-average reward.

In this study, the coefficient of variation (CV) of testing rewards was used as the main stability indicator. A configuration was considered stable when the CV of the testing rewards was below $10\%$, indicating low relative variability across the $50$ testing episodes.

\begin{table}[h]
\centering
\caption{Testing reward stability across selected traffic configurations.}
\label{tab:testing_reward_stability}
\begin{tabular}{lccccc}
\hline
\textbf{Scenario} & \textbf{Mean Reward} & \textbf{Std. Dev.} & \textbf{CV (\%)}  & \textbf{Stable?} \\
\hline
4 agents, sep=175  & 1146.962 & 74.694 & 6.512 &  Yes \\
6 agents, sep=150  & 1231.422 & 69.946 & 5.680 &  Yes \\
16 agents, sep=150 & 2363.857 & 81.233 & 3.436 &  Yes \\
18 agents, sep=150 & 2006.712 & 71.497 & 3.563 &  Yes \\
20 agents, sep=150 & 2454.078 & 87.355 & 3.560 & Yes \\
24 agents, sep=150 & 2188.306 & 86.455 & 3.951 &  Yes \\
\hline
\end{tabular}
\end{table}

As shown in Table~\ref{tab:testing_reward_stability}, all selected traffic configurations produced CV values below the $10\%$ stability threshold. The CV values ranged from $3.436\%$ to $6.512\%$, indicating low relative variability in testing reward across the $50$ randomized episodes. Therefore, the repeated testing results show that the fixed trained policies produced consistent reward performance when the total number of agents and the minimum-separation threshold remained the same, even though the initial aircraft positions and spacing varied across episodes. This testing stability supports the consistency of the learned policies under repeated randomized traffic arrangements. 

\section{Conclusion}
\label{sec:marl_conclusion_future_work}

This study developed a DQN-based MARL framework for decentralized conflict resolution among heterogeneous UAV and eVTOL aircraft operating under degraded surveillance conditions. The framework represented structured three-dimensional corridors, aircraft-specific motion and energy characteristics, lane-containment requirements, observation noise, communication delay, information dropout, wind disturbance, and model uncertainty. Separate policies were trained for UAV and eVTOL agents using local observations and a 14-action space that included maintain, turning, vertical, landing, and speed-control actions. The trained policies were evaluated across 90 combinations of aircraft density and minimum-separation threshold. LOS frequency increased as traffic density and separation requirements became larger, and higher-density cases also produced more events lasting several seconds. 

The action-selection results showed that the learned policies adapted to changing traffic conditions. When adequate separation was maintained, the maintain action accounted for approximately 79\% of decisions, whereas during LOS conditions, turning became the most frequent response at approximately 33\%, followed by maintain actions at 29\%, speed control at 25\%, and vertical maneuvers at 13\%. Turning produced a slightly greater average LOS-risk reduction than vertical maneuvering, and trajectory analysis showed that combined heading and speed changes could restore separation without requiring landing actions. Six Pareto-optimal configurations identified the best observed trade-offs between traffic capacity and LOS performance. In addition, the $2.76\%$ relative change in the 100-episode moving-average reward and testing reward coefficients of variation between $3.436\%$ and $6.512\%$ indicates stable learning and consistent simulated performance. Although these results do not provide certification-level evidence of real-world safety, the framework can support researchers, airspace planners, and traffic-management system developers in evaluating conflict-resolution strategies, corridor capacity, maneuver-priority rules, and degraded-surveillance procedures under different traffic and surveillance conditions.


\section*{Acknowledgment}

This research was funded by the Ohio Department of Transportation under Contract Number 136977 and Project ID Number 123151. The authors gratefully acknowledge this support.

\begin{spacing}{1}
\bibliography{sample}
\end{spacing}
\end{document}